\documentclass{article}

\usepackage{arxiv}

\usepackage[utf8]{inputenc} 
\usepackage[T1]{fontenc}    
\usepackage{hyperref}       
\usepackage{url}            
\usepackage{booktabs}       
\usepackage{amsfonts}       
\usepackage{nicefrac}       
\usepackage{microtype}      
\usepackage{lipsum}
\usepackage{graphicx}
\usepackage{amsmath}
\usepackage{amssymb}
\usepackage{mathtools}
\usepackage{multirow}

\title{Current Symmetry Group Equivariant Convolution Frameworks for Representation Learning}

\author{
 Ramzan Basheer \\
  Department of Avionics\\
  Indian Institute of Space Science and Technology- Thiruvananthapuram\\
  Kerala, India \\
  \texttt{ramzanbasheer.22@res.iist.ac.in} \\
   \And
 Deepak Mishra \\
  Department of Avionics\\
  Indian Institute of Space Science and Technology- Thiruvananthapuram\\
  Kerala, India \\
  \texttt{deepak.mishra@iist.ac.in} \\
}

\begin{document}
\maketitle
\begin{abstract}
Euclidean deep learning is often inadequate for addressing real-world signals where the representation space is irregular and curved with complex topologies. Interpreting the geometric properties of such feature spaces has become paramount in obtaining robust and compact feature representations that remain unaffected by nontrivial geometric transformations, which vanilla CNNs cannot effectively handle. Recognizing rotation, translation, permutation, or scale symmetries can lead to equivariance properties in the learned representations. This has led to notable advancements in computer vision and machine learning tasks under the framework of geometric deep learning, as compared to their invariant counterparts. In this report, we emphasize the importance of symmetry group equivariant deep learning models and their realization of convolution-like operations on graphs, 3D shapes, and non-Euclidean spaces by leveraging group theory and symmetry. We categorize them as regular, steerable, and PDE-based convolutions and thoroughly examine the inherent symmetries of their input spaces and ensuing representations. We also outline the mathematical link between group convolutions or message aggregation operations and the concept of equivariance. The report also highlights various datasets, their application scopes, limitations, and insightful observations on future directions to serve as a valuable reference and stimulate further research in this emerging discipline.
\end{abstract}

\section{Introduction}
Understanding the geometry of high-dimensional feature spaces in deep learning is crucial for designing robust and expressive models. These feature spaces often contain lower-dimensional manifolds that capture essential data relationships, enabling effective pattern learning and generalization \cite{cohen2019general}. 
Symmetries, fundamental to natural processes and a driving principle for new theories in physics, guide the generation of the data in question. Feature representations reflecting these symmetries disentangle data better than traditional methods. Network architectures designed with this principle achieve improved generalization and interpretability, forming the foundation of geometric deep learning \cite{bronstein2017geometric}. Its non-Euclidean treatment of data has garnered much interest among the machine learning community, as opposed to the traditional Euclidean way, considering only flat surfaces. A variety of methods are now available with improved statistical and computational efficiency compared to vanilla CNN methods.
\par
Group theory and its axioms describe symmetry mathematically. Equivariance with respect to the action of the symmetry group is the property that connects a feature map and the symmetry group of a neural network layer. This finding encourages us to design models with equivariant latent representation layers corresponding to the symmetries of interest as building blocks and the central constraint of our deep learning architectures \cite{villar2021scalars}. Such a broad framework has demonstrated tremendous potential for developing diverse architectures across various domains, each targeting distinct symmetry groups through different implementation techniques.
\par
\begin{figure}[ht]
  \centering
  \includegraphics[width=0.95\linewidth, keepaspectratio]{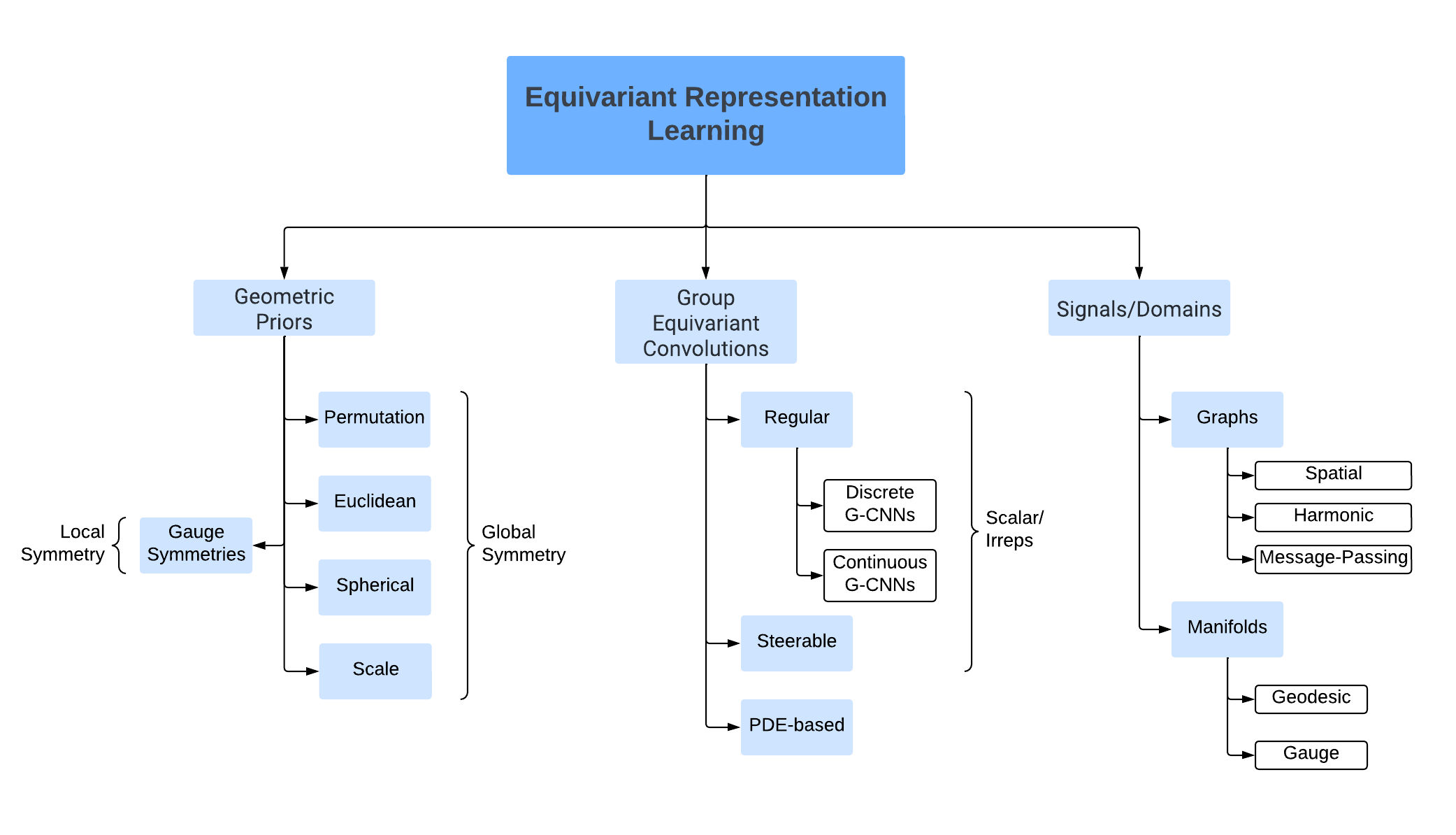}
  \caption{A taxonomy of symmetry group equivariant representation learning.}
  \label{fig:taxonomy}
\end{figure}

In this review, we focus on the mathematical connection between convolution operations and equivariant latent representations such that the models improve on size/complexity, interpretability, sample efficiency, generalizability, and stability. Among the many different geometric deep learning architectures, we consider only those that address equivariance as the geometric prior, omitting methods addressing invariance.  Figure \ref{fig:taxonomy} shows the complete taxonomy of current equivariant representation learning. Our contributions are summarized below:
\begin{itemize}
    \item We comprehensively cover the most recent theoretical developments and practical implementations of equivariant representation learning for graph and manifold data. We describe typical equivariant deep network design frameworks mathematically, using consistent notations that allow readers to use our work as a convenient reference.
    \item We categorize the equivariant convolutions into regular, steerable, and PDE-based group convolutions, and we further categorize them based on the domains they operate on and the target symmetries. These models vary based on domain, symmetry, and convolution type.
    \item We provide a list of benchmark datasets based on the tasks they target. Additionally, we explore the application scope of these methods in scientific and medical computations, offering insights into this research discipline.
    \item We discuss potential future directions that can steer this research, emphasizing the need for more abstract generalizations to study relationships between mathematical structures by exploiting their symmetry.
\end{itemize}

\textbf{Organization of the paper}: This paper outlines the diverse landscape of geometric deep learning methods based on their domain, symmetry, and feature aggregation approach. We begin with a background in geometric deep learning, the fundamental concept of equivariance, and representations in Section \ref{bckgrnd}. Subsequently, we investigate the domains where the data/signal resides in Section \ref{domain}. Our focus here is on graphs and manifolds, presenting prevalent equivariant convolutional techniques. In Section \ref{trichotomy}, we classify equivariant convolutions into regular, steerable, and PDE-based group convolutions. Section \ref{symmetry} discusses different models based on the symmetries they use as geometric priors. We cover equivariance to Euclidean symmetries, spherical symmetries, and the symmetries of general manifolds. We then list various datasets and discuss the scope of their applications in Section \ref{data_app}. We also outline the limitations of this discipline and the probable future scopes for addressing them in Section \ref{limit}. We summarise and conclude in Section \ref{conclude}. This review aims to serve as a valuable guide and reference for readers, particularly beginners, to equivariant representation learning. It aims to comprehensively guide their research endeavours and facilitate a thorough understanding of this highly sought-after machine learning technique and principle.

\section{Background}\label{bckgrnd}

\subsection{Geometric Deep Learning}
Geometric deep learning serves as an umbrella term for deep learning architectures that extend beyond regular grids, encompassing various data types and non-Euclidean domains. Previous methods in these domains included volumetric approaches \cite{maturana2015voxnet, riegler2017octnet, wang2017cnn} and multi-view strategies \cite{su2015multi, wang2018non, wang2019dominant}. PointNet \cite{qi2017pointnet} and PointNet++ \cite{qi2017pointnet++} operate directly on point cloud data but lack hierarchical structure, neglect global context, and incur higher computational and memory costs. They also remain susceptible to shape deformations and rigid body transformations. Models that leverage inherent geometric information within data offer potential solutions to these limitations. Consequently, practitioners of deep learning are increasingly drawn to these geometrically informed techniques over feature-based and vanilla convolutional methods. Geometric deep learning techniques revolve around treating signals as functions defined on certain domains \cite{bronstein2017geometric}. This categorization gives rise to two distinct problem classes within geometric learning: those that delineate the domain structure and those that analyze signals or functions defined on those domains. Many real-world problems involve identifying multiple domains, such as finding correspondences between two shapes. This often leads to the adoption of spatial convolutions as the most appropriate method for learning problems involving various domains and spectral methods for fixed domain problems. 
\par
The fundamental concept underpinning the generalization of deep learning models is the inductive bias \cite{mitchell1980need}, where prior knowledge or assumptions are baked into the model to guide its learning and preferred predictions or solutions. The most crucial geometric priors for a geometric deep learning model are stationarity and scale separation. The stationarity assumption can manifest as either invariance or equivariance with respect to specific geometric transformations. Invariant methods treat all input data points uniformly, resulting in a loss of information regarding the spatial relationships between points. They struggle to learn features sensitive to geometric transformations with no mechanism to explicitly capture the global context of the data. Equivariant convolutions, on the other hand, exploit inherent symmetries,
and can handle deformations, rotations, translations, and other data transformations. These convolutions adapt well to a wide range of non-Euclidean domains and transformations, making them more versatile.

\subsection{Equivariance}

A geometric deep-learning model operates based on three general concepts that significantly shape its functionality: {\itshape symmetry, representation, and equivariance}. The symmetry of a label function $\mathcal{L}:X \rightarrow Y$ on an input space ${X}$ to an output space ${Y}$ is a transformation $g:{X} \rightarrow {X}$ that leaves it unchanged. Mathematically this can be expressed using a function composition, denoted by the symbol $\circ$, as $\mathcal{L} \circ g = \mathcal{L}$. A priori knowledge about this label function's symmetry group, $G$, can make learning easy. If we know the label of an image, we can learn all the images obtained by applying transformations $g\in G$ to $x \in {X}$ that belong to the same class. The connection between $G$ and ${X}$ is a group action or {\itshape group representation}. A representation $\rho$ of the group $G$ is a {\itshape group action} of $G$ on a vector space $X$, $\rho:G \times X \rightarrow X$, by invertible linear maps such that 
\begin{equation}
    \rho(g)\rho(h) = \rho(gh), \, g,h \in G
\end{equation}
\par
Invariance to symmetric transformations yields limited statistical efficiency as the network would lose information about the relative spatial relationships between different features. This can negatively impact the network's ability to recognize and understand complex patterns in the data. Equivariance, by preserving relative pose of local features in the intermediate layers, can capture more detailed and informative data representations allowing subsequent layers to build upon these representations and make more accurate and meaningful predictions. Therefore, pursuing equivariance is imperative for optimal learning outcomes.
\par
\begin{figure}[ht]
  \centering
  \includegraphics[width=0.75\linewidth, keepaspectratio]{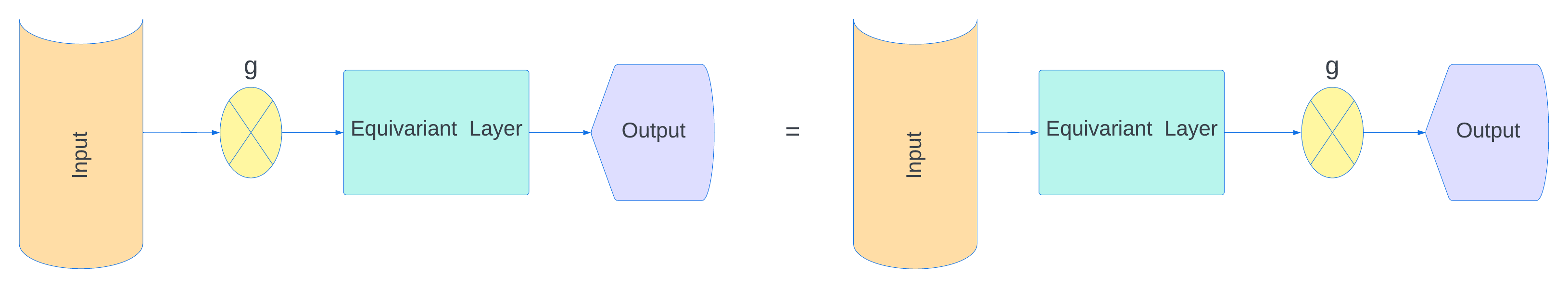}
  \caption{A block diagram showing equivariant action on a network layer.}
  \label{fig:eq_blk}
\end{figure}

A function $f: {X}\rightarrow{Y}$ is {\itshape equivariant} with respect to a group $G$ and group representations $\rho_g$ and $\rho_g^\prime$  in the input and output spaces respectively, if 
\begin{equation}\label{equivariance}
    f\left( \rho_g\left(x\right) \right)= \rho^\prime_g( f(x)), \, \forall g \in G
\end{equation}
Equation \ref{equivariance} signifies that for a given transformation of input with respect to a group action, a similar transformation acts on the output, maintaining the group structure, as illustrated in Figure \ref{fig:eq_blk}. These layers are then repeated leading to intermediate equivariant representations being functions on groups. The transformation $g$ on set of such feature maps $f \in F$ then is:
\begin{equation}\label{eqn:gonf}
    [\rho(g)f](x)=[f \circ g^{-1}](x)= f(g^{-1}x)
\end{equation}
\par
Equivariance is transitive: when each layer is equivariant, the whole network is equivariant. The input feature space extends into the group—a concept called lifting. Equivariance applies to convolutions $\star$ using filter banks $\psi$, shaping a multitude of geometric deep learning architectures. Mathematically an equivariant convolution between a function $f$ and filter bank $\psi$ has a general form as:
\begin{equation}\label{eqn:equivarConv}
    \left[\left[\rho(g)f\right] \star \psi\right](x) = \left[\rho(g)\left[f\star\psi\right]\right](x).
\end{equation}

\begin{figure}[ht]
  \centering
  \includegraphics[width=0.75\linewidth, keepaspectratio]{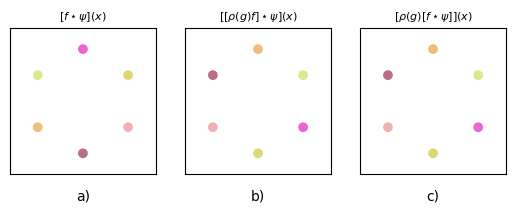}
  \caption{An example to show group equivariant convolution. a) Convolution output of inputs under no group action. b) Convolution output with respect to some group actions on input feature space. c) Convolution output with respect to group actions on output feature space}
  \label{fig:equivar}
\end{figure}
Equation \ref{eqn:equivarConv} indicates that the output of a convolution between a filter $\psi$ and rotated input feature $f$ equals the convolution followed by the rotation.  This is vividly illustrated through Figure \ref{fig:equivar}, depicting how the convolution output of transformed inputs matches the transformed output features. Determining the inherent symmetry within the data, acquiring an understanding of the symmetry through data-driven learning, or extracting symmetry information from the domain itself all emerge as pivotal steps in the process. We underscore the principle of equivariance in both features and learnable functions with respect to actions performed by a symmetry group identified within the data. Our focus here is solely on reviewing the connection between equivariant priors and convolutions while omitting discussions of equivariant nonlinearities, activations and normalization typically employed in such models.

\subsection{Steerable Representations}\label{steereps}

A {\itshape steerable representation} is composed of elementary {\itshape feature types} \cite{cohen2016steerable}. In {\itshape field theory}, a feature type corresponds to a specific field or physical quantity, encompassing scalar, vector, tensor, spinor, and gauge fields. Each feature type is associated with specific transformation properties or symmetry. Equivariant networks can generate steerable representations, enabling independent steering of each elementary feature.
\par
For the function $\mathcal{L}: \mathcal{X}\rightarrow\mathcal{Y}$, $\mathcal{Y}$ is steerable with respect to $G$, if the features $\mathcal{L} \circ f$ and $\mathcal{L} \circ \rho(g)f$ are related by a transformation $\rho^\prime(g)$ independent of $f$ for all transformations $g\in G$. This relationship is analogous to the one expressed in Equation \ref{equivariance}. The feature maps $f \in F$ obtained from the input can be decomposed into {\itshape fibers}. If $G$ is a group, and $H$ is its {\itshape stabilizer} group that fixes its origin, putting an equivariant constraint on filter banks $\Psi$ in $F$ with respect to $H$ results in $\rho(h)\Psi = \Psi \pi(h), \, \forall h \in H$. This leads to homomorphism of group representations $\pi, \rho$, referred to as {\itshape intertwiners}, $\text{Hom}_H(\pi,\rho)$. In other words, the output fibers are $H$-steerable by $\rho$ if the input space $F$ is $H$-steerable by $\pi$. Steerability can be achieved through {\itshape induced representation}, where the $H$-steerability of output fibers induces $G$-steerability of entire output feature space $F^\prime$. Let $\pi^\prime$ be a representation of $G$ induced by $\rho$ of $H$, i.e., $\pi^\prime = \text{Ind}_H^G \rho$. This leads to a mathematical expression of steerability as follows:
\begin{equation}\label{steerability}
    [\pi(g)f \star \psi](x) = [\pi^\prime(g)\psi \star f](x)
\end{equation}
While $\pi^\prime$ acts on entire output feature space $F^\prime$, $\rho$ acts only on individual fibers/vectors. The basic theory of equivariant representation learning for any data, including general manifolds, is encapsulated by principle bundle automorphisms \cite{cohen2019general}. In a broader sense, the symmetries of equivariant convolution networks are automorphisms $\alpha: P\rightarrow P$ acting on a principal bundle $P$ where $G$ acts on fibers transitively, effectively mapping fibers to fibers in an equivariant manner. 

\section{Domains of Equivariant Learning}\label{domain}

Geometric deep learning centers on leveraging low-dimensional structure in high-dimensional input spaces for efficient learning. This new paradigm considers data as signals on diverse domains, respecting its intrinsic geometry. Unlike traditional methods, it offers richer insights and better predictions. "5G"s in \cite{bronstein2017geometric} covers grids, groups, graphs, geodesics, and gauges as distinct domains of this new paradigm. We delve into various equivariant convolutions on graphs and manifolds, which aptly represent geometries of real-world non-Euclidean data.
\subsection{Graphs}
{\itshape Graph} $\mathcal{G}$ is a versatile widely adopted structure for modelling relationships between objects or entities in various real-world scenarios, such as computer networks, social networks, transportation systems, and many more. At its core, a graph comprises a collection of points (nodes or vertices, $\mathcal{V}$) connected by lines (edges $\mathcal{E}$). This relationship is denoted as $\mathcal{G = (V, E)}$. The vertices represent the objects, and the edges represent the relationships between them, which may be directed or undirected. In the realm of geometric deep learning, techniques aimed at graphs try to extend traditional convolution operations designed for regular grids to the irregular structure of graphs, using spatial, spectral, or spatio-spectral methods \cite{bronstein2017geometric}. Graph-based methods can be broadly categorized as invariant and equivariant Graph Neural Networks depending on their treatment of symmetries. In this review, we examine equivariant methods for processing graphs, a field rich for symmetry-preserving operations.
\par 
Keriven and Peyr\'e \cite{keriven2019universal} and Maron et al. \cite{maron2018invariant} characterized permutation equivariant graph neural networks, resulting in improved performance of such models. When the input is permuted, these models generate corresponding permutations in their output. A function $f: \mathbb{R}^{n^k} \rightarrow \mathbb{R}^{n^l}$ that processes a tensorized graph $\mathcal{G}$ is considered equivariant if $f(\sigma \star \mathcal{G}) = \sigma \star f(\mathcal{G})$. An equivariant graph neural network $f(\mathcal{G})$ is then characterized as 
    $f(\mathcal{G}) = \sum_s H_s \left [ \rho(F_s\left[ \mathcal{G}\right]+B_s)\right]+b$, 
where $F_s: \mathbb{R}^{n^d} \rightarrow \mathbb{R}^{n^{k_s}}$ and 
$H_s:  \mathbb{R}^{n^{k_s}} \rightarrow\mathbb{R}$ are linear equivariant functions, making the GNN globally equivariant, and $\rho$ is a pointwise non-linearity.  This general formulation can be easily extended to invariant graphs by choosing an invariant $H_s$. A simple proof for this extension is provided by Maehara and NT \cite{maehara2019simple}. It's worth noting that $k$-order GNNs can enhance the expressivity of these models while maintaining scalability \cite{maron2019provably, pan2022permutation}. 
\par
A message and update function are central to graph convolution \cite{gilmer2017neural, schutt2021equivariant}. Messages $\mathbf{m}_{ij}$ at layer $l$ are computed by $f_m$ from node embeddings $\mathbf{h}_i^l$ for a node $v_i \in \mathcal{V}$  and edge attributes $a_{ij}$ for $e_{ij} \in \mathcal{E}$ around its neighborhood $\mathcal{N}_{(i)}$.The edge attributes can provide context or additional features for the interaction between nodes, such as weight, distance, connection likelihood or some physical properties like conductivity, dipole moment, etc., that are meaningful for the task. These messages aggregated at a node are then forwarded to update the node embeddings of the subsequent layers by $f_u$.
\begin{gather}
    \mathbf{m}_i = \sum_{j\in \mathcal{N}_{(i)}}\mathbf{m}_{ij} =\sum_{j\in \mathcal{N}_{(i)}} f_m(\mathbf{h}_i^l, \mathbf{h}_j^l, a_{ij}) \label{eqn:MPN2}\\
    \mathbf{h}_i^{l+1} = f_u(\mathbf{h}_i^l, \mathbf{m}_i) \label{eqn:MPN3}
\end{gather}
Geometric embeddings can be injected into the network through positional coordinates $\mathbf{x}_i^l$ as in E(n) Equivariant Graph Neural Network (EGNN) \cite{satorras2021n}. The equations \ref{eqn:MPN2} and  \ref{eqn:MPN3} then become:
\begin{gather}
     \mathbf{m}_{ij} = f_m(\mathbf{h}_i^l, \mathbf{h}_j^l, ||\mathbf{x}_i^l - \mathbf{x}_j^l||^2, a_{ij}) \label{egnn2} \\
     \mathbf{x}_i^{l+1}=\mathbf{x}_i^l+C\sum_{j \neq i}(\mathbf{x}_i^l-\mathbf{x}_j^l)f_x(\mathbf{m}_{ij}) \label{egnn3}
\end{gather}
\par
A local context matrix through one-hot encoding of nodes and features captures neighborhood features and topology in message-passing frameworks \cite{vignac2020building}. Graph implicit functions with equivariant layers capture 3D details \cite{chen20223d}, yielding continuous 3D representations of various topologies. Directional embeddings lead to geometric message-passing \cite{gasteiger2021gemnet} using spherical filters $F_{sphere}$. Anisotropic kernels with topological flows yield stronger discriminative power \cite{beaini2021directional}. Equivariant subgraph aggregation improves over normal message-passing \cite{bevilacqua2021equivariant}. Combining geometric and topological data through high-dimensional simplex features, Eijkelboom et al. \cite{eijkelboom2023n} proposes $E(n)$ equivariant message passing networks (EMPSN).
\par
A harmonic approach decomposes graph functions into harmonics \cite{defferrard2019deepsphere}. Spherical message-passing disentangles physically-grounded representations \cite{liu2021spherical}. Spherical graph construction followed by graph convolutions encodes rotation equivariance \cite{yang2020rotation} of spherical images. The constraints and limitations of graph learning models regarding generalization and expressiveness are emphasized in various studies \cite{garg2020generalization, loukas2019graph, geerts2022expressiveness, keriven2021universality, alon2020bottleneck, balcilar2021analyzing, azizian2020expressive}.
\subsubsection*{Hyperbolic Graph}
While equations \ref{eqn:MPN2} and \ref{egnn2} can process non-Euclidean data, the node embeddings exist in Euclidean space, allowing to aggregate them using convenient metrics but potentially causing distortions.  Hyperbolic graph convolutions \cite{chami2019hyperbolic} embed nodes in hyperbolic space with negative curvature, resulting in smaller distortions. To achieve this, a hyperboloid manifold $\mathcal{H}^{d,K}$ of $d$ dimensions and constant negative curvature $-\frac{1}{K}, K>0$ is endowed with a Minkowski inner product.
Geodesics and distances are then defined on this hyperboloid manifold, enabling the mapping of vectors from the tangent space $\mathcal{T}_\mathbf{x}\mathcal{H}^{d,K}$ back to the manifold applying exponential and logarithmic maps evaluated at $\mathbf{x}$. In hyperbolic space, feature transform uses hyperboloid matrix multiplication $\otimes^K$ defined on $\mathcal{T}_\mathbf{x}\mathcal{H}^{d,K}$ \cite{chami2019hyperbolic} parameterized by a weight matrix $W^{d^\prime \times d}$. 
Initially, the hyperbolic points $\mathbf{x}^\mathcal{H}$ are projected onto $\mathcal{T}_\mathbf{x}\mathcal{H}^{d,K}$, followed by a transformation using $W^{d^\prime \times d}$, and ultimately projected back onto the manifold. The aggregation of messages will take place in the tangent space. Equations \ref{egnn3} and \ref{eqn:MPN3} in hyperbolic graph convolution will then be
\begin{equation}\label{eqn:HGCNagg}
    \text{AGG}^K(\mathbf{x}^\mathcal{H})_i = \exp_{\mathbf{x}^\mathcal{H}_i}^K\left(\sum_{j \in \mathcal{N}_{(i)}} w_{ij} \log_{\mathbf{x}^\mathcal{H}_i}^K\left(\mathbf{x}^\mathcal{H}_j \right) \right).
\end{equation}
\par
Instead of using approximate tangent spaces, Dai et al. \cite{dai2021hyperbolic} introduce aggregation and convolution operations on intermediate hyperbolic spaces. This approach involves a manifold-preserving hyperbolic feature transformation followed by hyperbolic aggregation as in \ref{eqn:HGCNagg}. This is achieved through a Lorentz linear transformation parameterized by matrices $W$ and $\widehat{W}$, where $\widehat{W}^T\widehat{W}=I$. In this process: $\bar{\mathbf{x}}_i^l$ is obtained by applying $W^l$ to $\mathbf{x}_i^{l-1}$.
    $\bar{\mathbf{x}}_i^{l,\mathcal{K}}$ is obtained by transforming $\bar{\mathbf{x}}i^{l,\mathcal{L}}$ using $p_{\mathcal{L \rightarrow K}}$.
    $\mathbf{m}_i^{l, \mathcal{K}}$ is calculated as the hyperbolic average on Klien model.
    $\mathbf{m}_i^{l, \mathcal{L}}$ is obtained by transforming $\bar{\mathbf{x}}i^{l,\mathcal{K}}$ using $p_{\mathcal{K \rightarrow L}}$. These transformations utilize concepts like Klien model representation, isometric and isomorphic bijections between Lorentz and Klien models, and others. Similar concepts explored by Lensink et al. \cite{lensink2022fully} with message propagation using hyperbolic equations and by Zhang et al. \cite{zhang2021lorentzian} with Lorentzian graph convolution show promise in various graph-based prediction tasks.
\subsection{Manifolds}
A {\itshape manifold} is a topological space that locally resembles Euclidean space. It provides a framework to represent complex data structures with intrinsic geometry. Such data, including geometric shapes, object surfaces, and point clouds, can possess irregular relationships. The goal of manifold learning is to uncover the intrinsic manifold structure within high-dimensional data and project it into a lower-dimensional space while preserving its fundamental geometric properties. Translation equivariant convolutions struggle to learn patterns from non-Euclidean data representations where object positions or shapes change. Extrinsic methods, like voxel-based and multi-view networks, \cite{wu20153d, su2015multi}, are memory-intensive and computationally expensive. Conversely, intrinsic methods focus on the manifold geometry and are robust to structural deformations and diffeomorphisms. 
\par
Spectral methods lack spatially localized filters for working on shapes and surfaces. Geodesic CNN \cite{masci2015geodesic} was the pioneering intrinsic CNNs on manifolds.  This introduced local geodesic polar coordinates, leading to topological discs $D(x)f$ that facilitate the extraction of local patches on the manifold using geodesic convolution:
\begin{equation}
    [f \star \psi](x) = \int_r \int_{\theta}\psi(t\theta,r)(D(x)f)(r,\theta)\, d\theta dr
\end{equation}
where $t\theta$ is a rotation by any arbitrary angle. Alternative operators, employing anisotropic heat kernels \cite{boscaini2016learning} as patches, removed radial dependencies, extending applicability to triangular meshes. A pseudo-coordinate system, along with a weight function, can extend CNNs to both graphs and manifolds \cite{monti2017geometric}. By jointly operating in the spatial-frequency domain, a convolutional construction can extend the Windowed Fourier Transform (WFT) on both manifold and point cloud data as an intrinsic operator \cite{boscaini2015learning}. PointNet \cite{qi2017pointnet} stands out as the first deep learning technique that works directly on the point cloud to learn the spatial coding of each point and aggregate them as a global feature, later enhanced by PontNet++ \cite{qi2017pointnet++}. Geometric perceptrons constrained by the geometry of neurons explore decision surfaces beyond hyperplanes. Exploring Klien geometries and Clifford Algebra \cite{melnyk2021embed, ruhe2023clifford}  show promise in this area.
\par
While earlier studies focused on the global symmetry of the manifold, gauge equivariant CNNs \cite{cohen2019gauge, he2021gauge, de2020gauge, basu2022equivariant,weiler2021coordinate} incorporate local symmetries. Since manifolds lack a canonical reference frame, gauges represent geometric quantities with respect to local reference frames. Beyond homogeneous manifolds, Di et al. \cite{di2022heterogeneous} introduced curvature-aware graph embeddings for heterogeneous manifolds. A natural progression involves the development of a network capturing tensorial curvature information. Ongoing research investigates how topological descriptors of manifolds and feature spaces influence learning across architectures, activations, and datasets and their impact on generalization and descriptive power.

\section{Group Equivariant Convolutions}\label{trichotomy}
Geometric deep learning aims to map feature spaces, effectively representing geometric quantities through convolutional operations while maintaining equivariance to specific symmetric transformation groups. Equivariant convolutions have proven instrumental, achieving high expressiveness alongside statistical and computational efficiency. From our studies, three convolutional operation categories for achieving equivariance emerge: regular group convolutions, steerable convolutions, and PDE-based convolutions. 

\subsection{Regular Group Convolutions}

A group convolution intuitively performs template matching. The template (kernel) is transformed and matched (inner product) under all possible transformations in a group, creating higher-dimensional feature maps as functions on the group.  The subsequent equivariant network layers perform similar template matching via group actions, effectively learning relative pose information hierarchically. These are linear operators with kernel constraints. Earlier works achieved only invariance to geometric transformations \cite{schutt2018schnet, gasteiger2020directional,liu2021spherical, coors2018spherenet}.  The EGNN \cite{satorras2021n}, formally characterized in equations \ref{eqn:MPN2} and \ref{egnn2}, stands as the pioneer equivariant model scalable to higher dimensions. These equations elucidate the pivotal role of scalar-vector multiplication in producing equivariant vectors, forming the core of these scalarization techniques. The scalar products and contractions involving scalar, vector, and tensor inputs adequately approximate an equivariant function \cite{villar2021scalars, gasteiger2021gemnet, geiger2022e3nn, jing2020learning}, bestowing them with universality.

\subsubsection{Discrete Setting}

Extending the mathematical framework of a standard translation equivariant convolution to a group convolution equivariant to any group $G$ is straightforward. In a typical convolutional network, the input feature map $f$ convolves with a filter bank $\psi$ as $\left[f \star \psi\right](x) = \sum_y \sum_k f_k(y) \psi_k(y-x)$. A translation operation $y \rightarrow y +t$, represented as $\rho(t)$ commutes with convolution, i.e., $[\rho(t)f \star \psi] = \rho(t)[f \star \psi](x)$, demonstrating translation equivariance. However, we cannot establish the same for other isometries of the grid structure. Given a group action $\rho(g)$ on a feature map $f$ at a point $x$, the transformed feature map can be obtained by finding the feature map that maps to $x$ under $g$, denoted as $g^{-1}x$ as in Equation \ref{eqn:gonf}. Cohen and Welling \cite{cohen2016group} proposed {\itshape group convolution} by substituting shift with elements of a transformation group $h \in G$, described mathematically as: 
\begin{equation}\label{grp_conv}
    [f \star \psi](g) = \sum_{h \in G} \sum_k f_k(h) \psi_k(g^{-1}h).
\end{equation}
\par
Here, the input feature space $\mathcal{X}$ is lifted to $G$ to generalize for operation between any network layer since the filters $\psi$ must also be functions on $G$. The concept of equivariance is vital to this characterization of convolutional neural networks derived from the basic principles of group theory. For an action $h\rightarrow uh$, we have $[\rho(u) f \star \psi](g) = \rho(u) [f \star \psi](g)$. Group convolution over hexagonal lattices \cite{hoogeboom2018hexaconv} enhances its effectiveness due to the hexagonal grid's greater degree of symmetry compared to a regular grid.
\par
An attentive group convolution operator $\star_\alpha$ represents another elegant extension of group convolution for emphasizing meaningful symmetry, defined in terms of attention map $\alpha$ that takes target and source elements $g, \tilde{g} \in G$ as $[f \star_\alpha \psi](g) = \sum_{h \in G} \sum_k \alpha(g, \tilde{g})f_k(h) \psi_k(g^{-1}h)$ \cite{romero2020attentive}. The Polar Transformer Networks (PTNs) combine the concepts of Spatial Transformer Networks and G-CNN to achieve equivariance \cite{esteves2017polar}, albeit limited to planar rotation groups. Other equivariant approaches include a network parameter-sharing scheme that induces a desirable symmetry group that acts discretely on feature maps of a network. Such a method is equivariant with respect to a group if the group induced can explain the symmetries of the network parameters \cite{ravanbakhsh2017equivariance}. Rotation Equivariant Vector Field Network (RotEqNet) \cite{marcos2017rotation} applies rotating convolutions at multiple orientations to return vector fields encoding maximum activation in magnitude and angle at every spatial location. Active Rotating Filters (ARFs) \cite{zhou2017oriented} actively rotate during convolution, where a virtual filter bank containing the filter and its rotated versions produces feature maps encoding location and orientation. Later extensions on G-CNN incorporate a more discrete group of symmetries. A cube group, $S_4$ of filters exhibiting cube-like symmetry, achieves equivariance to 3D right-angle rotations \cite{worrall2018cubenet}. $\mathcal{RCT}$-CNN adds a roto-scale-translation group for joint equivariance.

\subsubsection{Continuous Setting}
\subsubsection*{Compact Groups}

The group convolution in Equation \ref{grp_conv} works only for discrete groups and cannot be enumerated to large groups. Convolution on locally compact groups is an ideal generalization of this type. Integration with respect to the Haar measure $\mu$ defined on locally compact groups offers a way to meet this objective \cite{kondor2018generalization, finzi2020generalizing}. The Haar measure helps uniformly sample the continuous group. For a function $f$ and a filter $\psi$, both functions of a compact group, the convolution can be written as:
\begin{equation}\label{compact_conv}
    [f \star \psi](g) = \int_G f(h) \psi(g^{-1}h) \, d\mu(h)
\end{equation}
In cases where the feature space is the homogenous space of $G$, connection between the group $G$ and its homogenous space can be exploited. Group theory establishes that if $\mathcal{X}$ is a homogenous space on which some $g\in G$ acts transitively by fixing an origin $x_0 \in \mathcal{X}$ s.t. $x = g(x_0), x \in \mathcal{X}$, then the set of group elements that fix the origin $x_0$ form a subgroup $H$. The group elements that do this, $x_0 \mapsto x$, is called a {\itshape coset} and the set of all such cosets, $gH \coloneqq {gh|h \in H}$, is called {\itshape quotient space} $G/H$. The homogenous space $\mathcal{X}$ can be identified with $G/H$. A function in the homogenous space $f:G \rightarrow \mathbb{C}$ can be lifted to $G$ as $f\uparrow^G(g) = f(g(x_0))$. This connection between the group's structure and their homogenous spaces allows rewriting Equation \ref{compact_conv} as
    $\left[f \star \psi\right](g) = \int_G f\uparrow^G(h) \psi\uparrow^G(g^{-1}h) \, d\mu(h)$. \cite{xu2022unified} derived these homogenous spaces from a spectral sparsity-based kernel and non-linearity. 

\subsubsection*{Arbitrary Lie Groups}

While equation \ref{compact_conv} extends convolution to compact groups in a continuous context, practical formulation restricts us to finite compact groups. The convolution can be generalized to any arbitrary Lie group by employing a kernel expansion based on B-spline basis defined on the Lie algebra \cite{bekkers2019b}. The Schur-Poincar'{e} formula for the derivative of the exponential map permits Haar measure sampling for diverse Lie groups, thus achieving equivariance for such groups \cite{macdonald2022enabling}. Commonly, these methods consist of a lifting module, group convolution layers, and pooling layers, wherein the lifting module elevates the input feature space to the Lie group. Hutchinson et al. \cite{hutchinson2021lietransformer} extended this concept to a transformer architecture. Batatia et al. \cite{batatia2023general}, and Bogatskiy et al. \cite{bogatskiy2020lorentz} further extended it to reductive Lie groups such as the Lorentz group. The application of Riemannian Geometry on Lie groups can enhance spectral graph convolutions, yielding increased anisotropy \cite{aguettaz2021cheblienet} and ensuring equivariance on manifolds. The Lie group geometry can also be explored in Laplace distribution function space \cite{liaoLaplace}, facilitating the construction of equivariant layers. For a comprehensive list of popular architectures tailored to operate using regular group convolutions aligned with the prior symmetry group, refer to Table \ref{tab:RGCNN}.

\begin{table}[ht]
\centering
  \caption{Regular Group Equivariant Convolution Neural Networks}
  \label{tab:RGCNN}
  \begin{tabular}{lcc}
    \toprule
    Methods&Symmetry & Data type\\
    \midrule
    G-CNN  \cite{cohen2016group} &	Discrete Rotations & grids\\
    EGNN  \cite{satorras2021n} &	E($n$) & graphs\\
    GVP \cite{jing2020learning} &	E(3) & graphs\\
    GMN	 \cite{huang2022equivariant}	& E($n$)& graphs\\
    LieConv \cite{finzi2020generalizing}	& Lie Group & graphs\\
    PaiNN 	\cite{schutt2021equivariant}	& SO(3) & graphs\\
    LieTransformer	\cite{hutchinson2021lietransformer} &	Lie Group & point clouds/graphs\\
    TorchMD-NET \cite{tholke2022torchmd} &	O(3) & graphs\\
    GemNet \cite{gasteiger2021gemnet} &	SE(3) & graphs\\
    ChebLieNet	\cite{aguettaz2021cheblienet} &	SO(3) & grids/voxels\\
    Icosahedral CNN \cite{cohen2019gauge}&SO($n$) & grids/voxels\\
    B-Spline CNN \cite{bekkers2019b} &	Lie Group & grids \\
    CubeNet \cite{worrall2018cubenet} &	Cube Group & voxels\\
    RotEqNet \cite{marcos2017rotation} &	SO(3) & grids\\
    AGECN \cite{romero2020attentive} &	SE($n$) & grids\\
    SE(3)-Transformers \cite{fuchs2020se} & SO(3) & point clouds/graphs\\
    GESCNN \cite{kicanaoglu2019gauge}&SO(3)& grids/graphs\\
    SGCN \cite{yang2020rotation}&SO(3)& graphs\\
    DeepSphere \cite{defferrard2020deepsphere}&SO(3)& graphs\\
    E3NN \cite{geiger2022e3nn} &	E(3) & point clouds/graphs/voxels\\
    E-NF \cite{garcia2021n} &	E($n$ ) & graphs\\
    GEM-CNN \cite{de2020gauge}&SO(3)&graphs\\
    VN-DGCNN \cite{deng2021vector}& SO(3) & point clouds/voxels\\
    SE(3)-CNN \cite{kuipers2023regular} &	SE(3) & grids/voxels\\
    ScDCFNet \cite{zhu2022scalingtranslationequivariant} &	SE(3) & grids\\
    YOHO \cite{wang2022you} &	SO(3) & point clouds\\
    NequIP	\cite{Batzner2022} &	E(3) & graphs\\
    GET\cite{he2021gauge}&SO(3)&graphs\\
    SPConv	\cite{chen2021equivariant} &	SE(3) & point clouds/voxels\\
    VN-Tr \cite{assaad2022vn} & SO(3) & point clouds\\
    SFNO \cite{bonev2023spherical}&SO(3)&grids\\
    CGENN \cite{ruhe2023clifford} & E($n$) & point clouds \\
    GEVNet \cite{cortes2023higher}&SO($3$)&grids/voxels\\
    EMPSN \cite{eijkelboom2023n}&E($n$)&point clouds/graphs\\
  \bottomrule
\end{tabular}
\end{table}

\subsection{Steerable Group Convolutions}

Steerability in a feature space is achieved when the feature maps among all group elements rely on a signal-independent transformation. This property ensures the necessary equivariance to group actions and demonstrates significant successes in generalizing group convolutions to continuous symmetric groups. The pursuit of a steerable basis based on mathematical principles and particle physics theory has fuelled many exciting studies.

\subsubsection*{Irreducible Representations}

The representations of a compact group can be represented as the direct sum $\bigoplus$ of {\itshape irreducible representations} or {\itshape irreps} \cite{esteves2020theoretical}. This approach generalizes an equivariant model to $\text{SO}(3)$ \cite{dym2020universality}. Let $\rho_\mathcal{X}$ be a representation of $\text{SO}(3)$ on a vector space $\mathcal{X}$ (and $\rho_\mathcal{Y}$ on $\mathcal{Y}$) and $\rho(g)\vec{r}$ the action of $g \in \text{SO}(3)$ on $\vec{r} \in \mathbb{R}^3$ and $\rho_\mathcal{X}(g)x$ the action on $x \in \mathcal{X}$. The condition for {\itshape rotation equivariance} is $\mathcal{L}\circ \left[ \rho(g) \oplus \rho_\mathcal{X}(g) \right ] = \left [ \rho(g) \oplus \rho_\mathcal{Y}(g) \right] \circ \mathcal{L}$. We can transform the features into scalars, vectors, or tensors. The group elements, represented by $D^{(l)}$ called {\itshape Wigner D-matrices}, of $\text{SO}(3)$ are mapped to $(2l+1) \times (2l+1)$-dimensional matrices, where $l$ is the {\itshape rotation order}. The rotation order $l=0$ corresponds to scalars, $D^{(0)}(g)=1$, a trivial representation. $l=1$ corresponds to vectors, $D^{(1)}(g)=\rho(g)$. The higher-order representations can be either regular or decomposed as direct sums, $\bigoplus_l$, of irreducible representations via a change of basis ($Q$), $\rho(g) = Q^{-1} \left ( \bigoplus_l D^l(g) \right )Q$. Each block, i.e., Wigner-D matrix, only acts on a subspace $\mathcal{X}_{l_1} \subset \mathcal{X}$ which allows the factorization of the vector space $\mathcal{X} = \mathcal{X}_{l_1} \oplus \mathcal{X}_{l_2} \oplus \dots$. The $(2l+1)$-dimensional vector space on which $D^l$ acts will be called a {\itshape type} $l$ {\itshape steerable space}. Vectors of spherical harmonics are steerable by the Wigner-D matrices of the same degree, and steerable vectors represent the steerable functions on the groups $\text{S}^2$ and $\text{O}(3)$ \cite{brandstetter2021geometric}. Clebsch-Gordan (CG) coefficients $C^{lk} \in \mathbb{R}^{(2l+1) (2k+1) \times (2l+1)(2k+1)}$ provide an excellent means to decompose tensor products $\otimes$ of two irreps, facilitating the handling of higher-order tensors in geometric deep learning models.
\begin{equation}
    D^k(g) \otimes D^l(g) = (C^{lk})^{-1} \Biggl ( \bigoplus_{j= |k-l|}^{k+l} D^j(g) \Biggr )C^{lk}.
\end{equation}

\subsubsection*{Steerable Filters}

For a group-equivariant feature space $(F, \pi)$ and a network $\Phi: F \rightarrow F^\prime$, the output feature space $F^\prime$ is considered steerable with respect to $G$, if for all transformations $g \in G$ , the features $\Phi f$ and $\Phi\pi(g)f$ share a linear transformation $\pi^\prime(g)$ independent of $f$, $\Phi\pi(g) f= \pi^\prime(g)\Phi f$. Notable, $\pi^\prime$ is then a group representation induced by the representation $\rho$ of $H$, the stabilizer subgroup of $G$. The equivariant linear maps $\Phi$ between adjacent feature spaces are called intertwiners and have been well studied by Cohen et al. \cite{cohen2018intertwiners} for equivariant neural networks.
\par
Steerable Filter CNNs (SFCNNs), developed by Weiler et al. \cite{weiler2018learning}, efficiently compute orientation-dependent responses, enhancing sample complexity and promoting the generalization of learned patterns across orientations. A volumetric steerable approach, by Weiler et al. \cite{weiler20183d}, derives kernels analytically and parameterizes them using a complete steerable kernel basis. Convolution with $\text{SO}(3)$-steerable filters achieve equivariance between feature spaces, demonstrating remarkable accuracy and data efficiency in tasks like amino acid propensity prediction and protein structure classification. Cheng et al. \cite{cheng2018rotdcf} adopts a similar decomposition strategy that combines steerable bases and group geometry to create a Rotation-equivariant CNN with decomposed convolutional filters (RotDCF). This concept is extended by Graham et al. \cite{graham2020dense} to histology image analysis for computational pathology tasks.

\subsubsection*{G-Steerable kernels}

Steerable settings impose constraints on convolutional kernels leading to a general characterization of $G$-steerable kernel spaces. Weiler and Cesa \cite{weiler2019general} simplify these constraints under irreducible representations, thereby extending their applicability to arbitrary group actions. Through the generalization of the Wigner-Eckart theorem for spherical tensor operators, Lang and Weiler \cite{lang2020wigner} established the characterization of the steerable kernel space for any compact group $G$. This was accomplished in terms of generalized reduced matrix elements, Clebsch-Gordan coefficients, and harmonic basis functions on homogeneous spaces. Shen et al. \cite{shen2022pdo} introduced PDO-s3DCNNs, a general steerable 3D-CNN utilizing Partial Differential Operators (PDOs) equivariant to $\text{SO}(3)$ and its common discrete subgroups. Zhdanov et al. \cite{zhdanov2022implicit} presents a kernel parameterization employing an MLP satisfying equivariance constraints, generalizable to arbitrary groups $G \leq O(n)$, for constructing an implicit representation of steerable kernels for compact groups. Recent advancements, like the theory of 3D steerable spherical neurons proposed by Melnyk et al. \cite{melnyk2021embed}, \cite{melnyk2022steerable, melnyk2023deep}, explore the construction of equivariant hyperspheres for point-cloud analysis. Novel approaches delve into alternative algebraic spaces for representations, such as the Clifford algebra-based equivariant parameterization \cite{ruhe2023clifford}. Additionally, Zhu et al. \cite{Zhu_2023_CVPR} augments the KPConv \cite{thomas2019KPConv} with steerability constraints to enable steerable convolutions on point clouds represented in a quotient space. The symmetrical priors and data types of popular steerable architectures are summarized in Table \ref{tab:SGCNN}.

\begin{table}[ht]
\centering
  \caption{Steerable Group Convolution Neural Networks}
  \label{tab:SGCNN}
  \begin{tabular}{lcc}
    \toprule
    Methods&Symmetry & Data type\\
    \midrule
    TFN  \cite{thomas2018tensor} &	SE(3) & point clouds\\
    Cormorant  \cite{anderson2019cormorant} &	SO(3) & graphs \\
    SE(3)-Tr  \cite{fuchs2020se} &	SE(3) & point clouds/graphs \\
    SEGNN  \cite{brandstetter2021geometric} &	E(3) & graphs \\
    SESN  \cite{sosnovik2019scale} &	scale & grids \\
    SE3CNN  \cite{weiler20183d} &	SE(3) & voxels \\
    SFCNN  \cite{weiler2018learning} &	SE(2) & grids \\
    3DSCNN  \cite{weiler20183d} &	SE(3) & voxels \\
    RotDCF  \cite{cheng2018rotdcf} &	SO(2) & grids \\
    DSF-CNN  \cite{graham2020dense} &	SE(2) & grids \\
    E2SCNN  \cite{weiler2019general} &	E(2) & grids \\
    PDO-s3DCNN  \cite{shen2022pdo} &	SO(3) & voxels \\
    SESN  \cite{sosnovik2019scale} &	scale & grids \\
    E2PN  \cite{Zhu_2023_CVPR} &	SE(3) & point clouds\\
    RSESF  \cite{yang2023rotation} &	scale & grids \\
    SECNN  \cite{wimmer2023scale} &	scale & voxels \\
  \bottomrule
\end{tabular}
\end{table}

\subsection{PDE-based Group Convolutions}

PDE-based G-CNNs are the latest framework in the geometric deep learning ecosystem proposed by Smets et al. \cite{Smets2023} as a parameterization of the equivariant layer weights with PDE-coefficients encoding the geometry. The idea is a straight extension of PDE-based image processing on Euclidean spaces to a more general homogenous space $\mathcal{M} = G/H$ where $H$ is a subgroup of $G$. 
The blueprint of a PDE-G-CNN is a little different from the blueprint postulated by Bronstein et al. \cite{bronstein2017geometric} for a general geometric deep learning architecture. Since the input and output may not necessarily belong to homogenous spaces, the input is lifted to $G/H$ space before it is passed to the PDE layer and projected before it is drawn as output from the PDE layer. We only focus on PDE layer design and leave the interested readers to refer \cite{bekkers2018roto} for lifting and \cite{smets2019geometric} for projecting layer designs.
\par
Each  PDE layer operates as a PDE solver of a set of time-stepping equations with inputs as its initial conditions. The solutions obtained after a set time interval undergo an affine transform to form the outputs or inputs to the subsequent PDE layer. The action of $l$-th PDE layer on a function $U_l \in G/H$ by the operator $\nabla_{T, \theta}, T \geq 0$ parameterized by $\theta_l$ with learnable affine transform coefficients $a_l$ and $b_l$ is
\begin{equation}\label{eqn:PDE}
    U_{l+1}= \sum_i a_{l,i} \nabla_{T,\theta_{li}}(U_l) + b_l.
\end{equation}
The final approximate solution will have four terms: convection, diffusion, dilation and erosion, replacing the smoothing, shifting, max-pooling, and min-pooling layers of a vanilla CNN architecture.
\par
Bellaard et al. \cite{Bellaard2023, bellaard2023geometric} proposes a sub-Riemannian distance approximation for defining the morphological convolution kernel of PDE-G-CNN. Pai et al. \cite{pai2023functional} demonstrate the geometric interpretability of PDE-G-CNNs through theoretical underpinnings of their equivariance while also providing experimental evidence that showcases improved statistical efficiency. Helwig et al. \cite{helwig2023group} apply PDE solvers for building equivariant networks operating in the frequency domain leveraging the equivariance of Fourier transform. The innovative PDE-G-CNN framework represents a cutting-edge advancement in the realm of group equivariant convolutions. Its profound potential for addressing significant research gaps within the domain of geometric learning stands poised to make substantial and lasting contributions in the forthcoming years.

\section{Geometric Priors} \label{symmetry}

We have understood what role symmetry plays in equivariant representation learning. Selecting an appropriate symmetry for the task at hand determines the inductive bias of the model and its computational cost. The choice of symmetry refers to the transformations that the model can identify and preserve within its learned representations. These transformations include translations, rotations, reflections, permutations, scale, gauge transformations, or combinations thereof. The translation, rotation, and reflection symmetries can be grouped under Euclidean symmetries, whereas gauge transformations are under the symmetries of a general manifold. Permutation and scale symmetries are treated as discrete symmetry groups; we will address each of them individually. This section reviews the connection between convolutions and equivariance in these transformations. The choice of symmetry shapes the model's inductive bias and computational complexity for the task. This choice dictates which transformations the model can identify and retain within its learned representations. 

\subsection{Permutation Symmetries}

Early works characterized permutation-equivariant objective functions \cite{zaheer2017deep, maron2018invariant, qi2017pointnet} on unordered sets and point clouds with limited universal approximation. Approximation with polynomial characterization \cite{segol2019universal, sannai2019universal, keriven2019universal} expands the scope to graph data, achieving permutation-equivariant universal approximation. The expressive power of graphs through structural message-passing leverages permutation equivariance \cite{vignac2020building}. General permutation-equivariant networks \cite{thiede2020general, de2020natural, azizian2020expressive, sun2020acne} exhibit effectiveness in graph learning tasks like link prediction and molecular generation, exploiting the geometric quantities of the graph data. Enhancing computational efficiency involves parameterizing and making permutation-equivariant layers sparsity-aware \cite{godfrey2023fast, morris2022speqnets}. Recent developments employ parameter sharing scheme \cite{zhou2023permutation}and variational auto-encoders generative models  \cite{hy2023multiresolution} to build learnable permutation equivariant layers, harnessing the permutation equivariance of unordered hidden neurons in a deep network. Embracing hierarchy and stochasticity of latents leads to more expressive equivariant graph models \cite{joshi2023expressive}.

\subsection{Euclidean Symmetries}

In the context of 3D space, the symmetry group encompasses translations, rotations, and reflections. These elements allow us to transition between coordinate systems and predict how quantities change under such transformations. The studies below explore the integration of these symmetries into the group convolution of equation \ref{grp_conv}. Rotational symmetry preserves the geometric properties of an object rotated by some angle. Formally, it is a symmetry with respect to rotations in $n$-dimensional Euclidean space, making it an $E(n)$ subgroup.  Since rotational symmetry around all points suggests translational symmetry, we consider rotational symmetry about a point or origin, forming the special orthogonal group $\text{SO}(n)$. Most studies discussed here consider $n=3$, i.e., the rotation group $\text{SO}(3)$ which is a Lie group, and differs in their particular convolution operation whether spatial, spectral or message-passing. 

\subsubsection*{Spatial Convolutions}

A rotation-equivariant network consists of a stack of rotation-equivariant convolutional layers that establish a connection between rotations in the input image and rotations in the feature map produced for each filter. Dieleman et al. \cite{pmlr-v48-dieleman16} combine four cyclic operations cast as layers to make a network partially equivariant to rotations. Cohen and Welling \cite{cohen2016group} incorporated a 4-fold rotation equivariance into CNNs through {\itshape group convolutions}, mathematically formulated as in equation \ref{grp_conv}, achieving far greater weight sharing than regular convolution layers. A G-CNN filter recognizes relative pose charateristics and can match such feature constellations in every global pose. Equivariance to symmetry transformations, besides improving statistical efficiency, can also facilitate geometric reasoning. A complimentary group convolution involving primary and secondary group convolutions can reduce the parameter and computation complexity \cite{zhang2017IGCNet}. The interleaving of these complementary convolutions generalizes regular and group convolutions.
\par
While these papers deal with $\text{SO}(2)$, \cite{kondor2018generalization, cohen2019general, finzi2020generalizing} generalized convolutions to the actions of compact groups, homogenous spaces, and Lie groups, respectively. Esteves \cite{esteves2020theoretical}  discusses the theoretical aspects of G-CNNs based on these works. Unlike invariant multiple-view aggregations, Esteves et al. \cite{esteves2019equivariant} proposed group convolution over an icosahedral group, enabling joint reasoning using only a few input polar view representations and guaranteeing equivariance. Further to this representation, a cylindrical canonical representation extended the rotation equivariance of PTNs \cite{esteves2017polar} to a 3D setting. 
\par
$\text{SE}(3)$ separable convolution promises to lower the computational cost of convolutions between 6D functions in $\text{SE}(3)$ space without compromising the expressiveness of the equivariant features \cite{chen2021equivariant}. This is achieved by separating the convolution into two operators, one in 3D Euclidean space and another in $\text{SO}(3)$ space. The $\text{SE}(3)$ space is first discretized  into finite 3D spatial points $X:\{x|x\in \mathbb{R}^3\}$ and the group $G \subset \text{SO}(3)$. For a feature mapping $f(x_i, g_j):X \times G \rightarrow \mathbb{R}^D$, the discrete convolution is
\begin{equation}
\begin{aligned}
    \left [ f \star h\right](x,g) & =  \sum_{x_i \in X}\sum_{g_j \in G}f\left(g^{-1}\left(x-x_i\right),g_j^{-1}g\right)h(x_i, g_j) \\
    & =  \sum_{x_i^\prime \in X_g}\sum_{g_j \in G}f(x-x_i^\prime,g_j^{-1}g)h(gx_i^{-1}, g_j)
\end{aligned}
\end{equation}
where $X_g:\{g^{-1}x|x \in X \}$ is the rotated set, $X \times G$. is the domain of the convolution kernel $h$ with size $ |X| \times |G|$. $h$ can now be separated into two kernels $h_1$ and $h_2$ with kernels sizes $|X| \times 1$ and $1 \times |G|$ belonging to smaller domains $X \times \{\mathbf{I}\}$ and $\{\mathbf{0}\} \times G$ respectively. The split convolution is
\begin{equation}
\begin{aligned}
    \left[f \star h_1\right](x,g) & =  \sum_{x_i^\prime \in X_g}f(x-x_i^\prime,g)h_1(gx_i^\prime, \mathbf{I}) \\
    \left[f \star h_2\right](x,g) & =  \sum_{g_j \in G}f(x,g_j^{-1}g )h_2( \mathbf{0}, g_j) .
\end{aligned}
\end{equation}
\par
Knigge et al. \cite{knigge2022exploiting} propose a continuous separable convolution through lifting and group convolutions, but only in the  $\mathbb{R}^2$ domain. Recently, Kuipers and Bekkers \cite{kuipers2023regular} devised a separable group convolution on volumetric data by splitting the kernel in continuous rotation and spatial domain. The continuous $\text{SO}(3)$ kernels are parameterized using radial basis functions (RBFs) to approximate the continuous group integral. The $\text{SE}(3)$ convolutions are defined on the semidirect group $\mathbb{R}^3 \rtimes \text{SO}(3)$. Its separated kernel is $k_{\text{SE}(3)}(\mathbf{x, R}) = k_{\text{SO}(3)}(\mathbf{R})k_{\mathbb{R}^3}(\mathbf{x})$. The continuous ${\text{SO}(3)}$ integral is first discretized to form uniform ${\text{SO}(3)}$ grids from where the continuous ${\text{SO}(3)}$ kernels are interpolated using RBF interpolation. Using Gaussian RBF $\psi$ and Riemannian distance $d$, RBF interpolation coefficient $a_{d, \psi}$ can be defined for $\mathbf{R}$ corresponding to each grid element $\mathbf{R}_i$ of resolution $N$. The continuous kernel, then, is $k_{\text{SO}(3)}(\mathbf{R}) = \sum_{i=1}^N a_{d, \psi}(\mathbf{R}, \mathbf{R}_i)\mathbf{k}_i $.
\par
Extending 1D scalars to 3D vectors, Deng et al. \cite{deng2021vector} used Vector Neuron representations to map $\text{SO}(3)$ actions to latent space. The scalar neuron representation $z \in \mathbb{R}$ is lifted to a vector representation $v \in \mathbb{R}^3$. A linear map $f_{lin}(\cdot;\cdot)$ is built acting on these Vector Neurons and is verified to commute with $\mathbf{R} \in \text{SO}(3)$ for equivariance
\begin{equation}
    f_{lin}({V};{W})={WV}
    f_{lin}({V}\mathbf{R}; {W})={WV}R\mathbf{R} f_{lin}({V}; {W})\mathbf{R}
\end{equation}
A transformer architecture from vector neurons called VN-Transformer \cite{assaad2022vn} eliminated the feature processing of vector neurons \cite{deng2021vector} and supported non-spatial attributes. 
\par
Other representations have also been tried for equivariant representation learning lately. Shen et al. \cite{shen20203d} showed that when feature representations are in quaternions, the network has rotation-equivariance property naturally besides permutation invariance. A $u$-th point $([x_u, y_u, z_u]^T)\in \mathbb{R}^3$ in a 3D data can be represented as a quaternion $\mathbf{x}_u = 0 + x_u \mathbf{i} + y_u \mathbf{j} + z_u \mathbf{k}$. Rotating it around axis $\mathbf{o}= 0+ o_1 \mathbf{i} + o_2 \mathbf{j} + 03 \mathbf{k}$ (where $o_1, o_2, o_3 \in \mathbb{R}$, $||o||=1$), with an angle $\theta \in [0, 2\pi)$ to get $\mathbf{x}_u^\prime$, the rotation by  a unit quaternion $\mathbf{R}= \cos \frac{\theta}{2}+ \sin \frac{\theta}{2}( o_1 \mathbf{i} + o_2 \mathbf{j} + 03 \mathbf{k}) = e^{\mathbf{o}\frac{\theta}{2}}$ and its conjugate $\overline{\mathbf{R}}$ can be represented as 
    $\mathbf{x}_u^\prime = \mathbf{R}\mathbf{x}_u\overline{\mathbf{R}}$.
The rotation equivariance is then 
    $f({x}^{(\theta)})=\mathbf{R} \circ f({x}) \circ \overline{\mathbf{R}} \; \;\;\text{s.t.}\;\;\; {x}^{(\theta)}\triangleq \mathbf{R} \circ {x} \circ \overline{\mathbf{R}}$.
Later an explicit latent rotation parametrization is used in quaternion equivariant capsule networks \cite{zhao2020quaternion} for equivariance through a dynamic routing algorithm on quaternions. \par
It's essential to point out that G-CNNs \cite{cohen2016group} apply only to discrete groups and are infeasible for larger groups. Brandstetter et al. \cite{brandstetter2021geometric} demonstrated the effectiveness of steerable equivariant message-passing networks (SEGNNs) in encoding covariant information, encompassing scalars, vectors, and tensors within node and edge attributes. This enables the processing of higher-order geometric quantities. However, acquiring an alternative basis and obtaining Clebsch-Gordan coefficients is not a straightforward task, and they yield nontrivial representations that operate within those algebraic subspaces. Ruhe et al. \cite{ruhe2023clifford} incorporate equivariant actions into the Clifford algebraic space to form Clifford Group Equivariant Neural Networks (CGENN), bypassing the requirement for an equivariant alternative basis. However, the kernel expansion in a steerable setting and the parameterisation technique of CGENN could result in computational overhead. Symmetry-breaking techniques \cite{passaro2023reducing} could potentially alleviate this inefficiency. Dym and Maron \cite{dym2020universality} studied the approximation and expressive powers or rotation equivariant networks by characterising the space of equivariant polynomials.
\par
Hoogeboom et al. \cite{hoogeboom2018hexaconv} showed how hexagonal lattice symmetry could be exploited to achieve this. Although Bekkers et al. \cite{bekkers2018roto}  and Lafarge et al. \cite{LAFARGE2021101849} proposed  $\text{SE}(2)$ convolution frameworks, we will be covering only 3D rotations and translations here. Winkels and Cohen \cite{winkels20183d} extended the G-CNN \cite{cohen2016group}  to three dimensions by incorporating 3D roto-reflection groups where filters are cubes instead of squares. The roto-translation equivariant model by Worrall and Brostow \cite{worrall2018cubenet} exhibited equivariance to translations and rotations by right angles in three dimensions. Equivariance to cube-like symmetry is achieved by considering the set of all right-angle rotations of a cubic filter $\mathbf{F}_x \in \mathbb{R}^{N \times N \times N}$ to form a {\itshape cube group} $S_4$ with 24 such rotations. The Klein's four-group $V$ is a subgroup of $S_4$. Roto-translation, a transformation of rotation followed by translation, can be composed as $tr$ for $t\in \mathbb{Z}^3$ and $r \in V$. The rotation and translation will result in permutations in convolution space which can be figured out from {\itshape Cayley Table}. The entire process is mathematically formulated as:
\begin{equation}
     \left[f\star \psi \right]_{tr}=\sum_{\tau \in \mathbb{Z}^3}\sum_{\rho\in V}\left[tr \psi \right]_{\tau\rho}\left[f\right]_{\tau\rho}=\sum_{\tau \in \mathbb{Z}^3}\sum_{\rho\in V}\left [t\left[r\psi \right]_\rho \right]_\tau f_{\tau\rho}
\end{equation}
The equivariance of the group convolution between filter $\psi$ and feature map $f$ is verified from the following relation for a transformation $p$ and permutation matrix $P_p$
\begin{equation}
    \left[pf\right]\star \psi = p\left[f\star \psi\right]=P_p \left[f\star \psi\right]
\end{equation}

\subsubsection*{Harmonic Convolutions}

Worrall et al. \cite{worrall2017harmonic} replaced regular CNNs with circular harmonics that are locally equivariant to continuous $360^\circ$ by restricting the filters to be of circular harmonics ${W}_m(r,\phi;R,\beta)=R(r)e^{i(m\phi+\beta, )}$, where $(r,\phi$) is the image/feature maps spatial coordinates, $m$ is rotation order, $R$ is the filter radial profile, and $\beta$ is the filter orientation-selectivity. The rotation equivariance for a $\theta$-rotated image $f^\theta$ with respect to unrotated image $f^0$ is verified from the relation
\begin{equation}
    [{W}_m \star f^\theta]=e^{im\theta}[{W}_m \star f^0]
\end{equation}Every receptive field patch thus returned maximal response and orientation. 

\par
The Tensor Field Networks (TFN) introduced by Thomas et al. \cite{thomas2018tensor} has layers that process geometric inputs and outputs as scalars, vectors, and higher-order tensors using filters embedded in spherical harmonics. These spherical harmonics are functions mapping from a sphere to complex or real numbers, and they exhibit equivariance to $\text{SO}(3)$, similar to the utilization of circular harmonics \cite{worrall2017harmonic}. We obtain the tensor product of two irreducible representations of different orders with $u^{(l_1)} \in l_1$ and $v^{(l_2)} \in l_2$ using the equivariant property of the tensor product and Clebsch-Gordan coefficients \cite{griffiths2018introduction}
\begin{equation}
    (u \otimes v)_m^{(l)} = \sum_{m_1 = -l_1}^{l_1}\sum_{m_2 = -l_2}^{l_2} C^{(l,m)}_{(l_1, m_1), (l_2, m_2)}u^{(l_1)}_{m_1}v^{(l_2)}_{m_2} 
\end{equation}
Point convolution, as demonstrated in Sch\"{u}tt et al. \cite{schutt2017schnet}, guarantees equivariance owing to the equivariance of both the filters and the Clebsch-Gordan coefficients. {\itshape Irreps} are once again used in N-body networks by Kondor \cite{kondor2018n} where the output states of subsystem neurons transform under rotations according to the irreducible representations of the rotation group, thus achieving equivariance. Anderson et al. \cite{anderson2019cormorant} also represented all activations in the form of spherical tensors with non-linearity introduced through the Clebsch-Gordan transform to construct Covariant Molecular Neural Networks (Cormorant). The channel-wise CG-product involving $F_{l_1} \in \mathbb{C}^{(2l_1+1)\times n_1} $ and $G_{l_2} \in \mathbb{C}^{(2l_2+1)\times n_2} $ is akin to the approaches used in Kondor et al. \cite{kondor2018clebsch} and Thomas et al. \cite{thomas2018tensor} expressed as:
\begin{equation}
    \left[ \left[ F_{l_1}\otimes_{cg}G_{l_2}\right]_l \right]_{\ast, i}=C_{l_1, l_2, l}\left(\left[F_{l_1}\right]_{\ast,i}\otimes \left[G_{l_2} \right]_{\ast,i}\right).
\end{equation}
\par
$\text{SE}(3)$-Transformer introduced by Fuchs et al. \cite{fuchs2020se} adds attention weights to the edge-wise equivariant messages calculated using convolutions from TFN \cite{thomas2018tensor} and a self-interaction layer, formally expressed for a neighbourhood $\mathcal{N}_i \subset \{1, \dots , N\} $, learnable weight kernel $W_V^{lk}:\mathbb{R}^3 \rightarrow \mathbb{R}^{(2l+1) \times (2k+1)} $ expressed in $V$-th spherical harmonic, and edge-wise attention weights $\alpha_{ij}$ as $f_{out,i}^l = W_V^{ll}f_{in,i}^l + \sum_{k \geq 0}\sum_{j \in \mathcal{N}_i} \alpha_{ij}W_V^{lk}(x_j - x_i) f_{in,j}^k$. An iterative version of this method outperformed the single-pass version \cite{fuchs2021iterative}, demonstrating that an iterative model can yield benefits in certain learning tasks.
\par
Recently Zhu et al. \cite{zhu2023e2pn} proposed Efficient $\text{SE}(3)$ Equivariant Point Network (E2PN) for processing point cloud data with equivariance up to a discretization on $\text{SO}(3)$. The $\text{SE}(3)$ feature maps are reduced to $\text{S}^2 \times \mathbb{R}^3$ where $\text{S}^2 = \text{SO}(3)/\text{SO}(2)$ is the {\itshape quotient space} of $\text{SO}(3)$ given the stabilizer group $\text{SO}(2)$. An $\text{E}3 \times \text{SO}(3)$ equivariant network proposed by Elaldi et al. \cite{elaldi20233} constructs equivariant sparse convolution layers for voxelized signals manifested as spheres. Recent generative models equivariant to Euclidean symmetries have outperformed the baseline models showing their strength to lift the features from discrete to continuous space \cite{garcia2021n}.

\subsection{Spherical Symmetries}

We distinguish spherical symmetries from Euclidean ones by adopting the surface of a sphere as the underlying geometry. Such a geometric consideration is crucial in applications including weather modelling, omnidirectional views of autonomous vehicles, drones, and robots, molecular regression problems, astrophysics, chemistry and  more. In these scenarios, spherical correlations can be defined appropriately to exploit the spherical symmetry of the signal domain. We investigate some of the most successful strategies and their evolution, which either lift the inputs to $\text{SO}(3)$ functions or apply convolutions directly on the sphere.

\begin{figure}[ht]
  \centering
  \includegraphics[width=0.75\linewidth, height=0.2\paperheight, keepaspectratio]{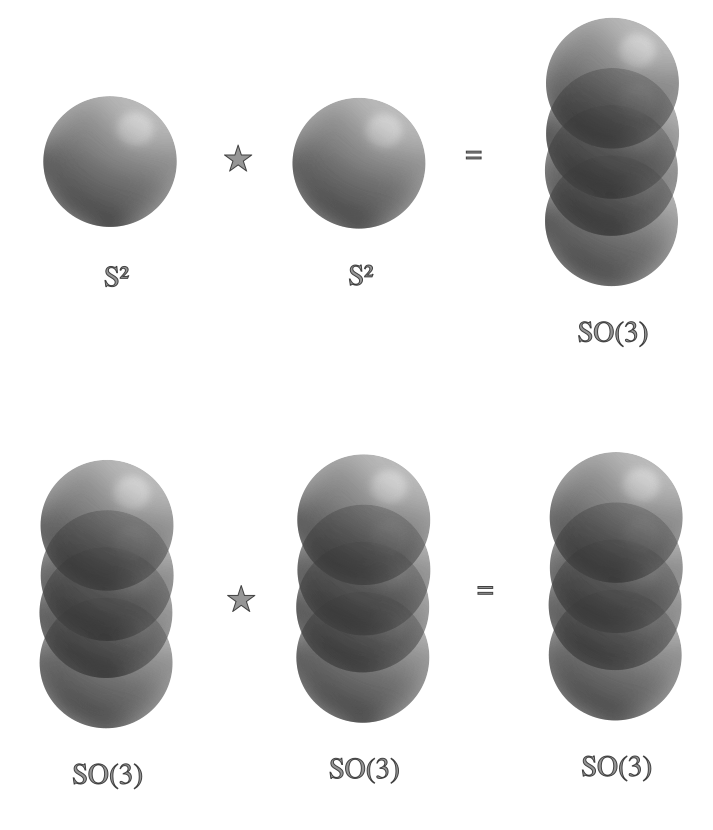}
  \caption{Spherical convolution}
  \label{fig:sphericalConv}
\end{figure}

\subsubsection{Spatial Convolutions}

We cannot trivially extend the convenience of translational convolution to a spherical signal as weight sharing is not an isomorphism in a sphere $\text{S}^2$. Addressing this limitation, Boomsma and Frellsen \cite{boomsma2017spherical} introduced two strategies for realizing convolutions on spherical volumes through discretization on a sphere. First, using a spherical-polar grid and next, a cubed-sphere transformation. As we move from the equator to the poles in the spherical grid strategy, the spacing between grid points decreases, reducing the weight-sharing capacity of the filters between different areas of the sphere. The cubed-sphere transformation is an improvement upon this limitation. It achieves a more regular grid by decomposing the sphere into six patches, allowing conventional convolution to be applied with the added benefit of using tensors for representing geometric features, though with certain artefacts at the corners. Coors et al. \cite{coors2018spherenet} also had a similar uniform sphere sampling for learning spherical image classification and object detection. Fox et al. \cite{fox2022concentric} suggested a new concentric spherical spatial grid structure to build a Concentric Spherical Neural Network (CSNN). 

\subsubsection{Graph Convolutions}

Khasanova and Frossard \cite{khasanova2017graph} proposed graph-based representations and graph convolution for omnidirectional/spherical images. A longitude $\theta_k \in [-\pi, \pi]$ and latitude $\phi_k \in [-\frac{\pi}{2}, \frac{\pi}{2}]$ uniquely identifies a point $\mathbf{x}_k$ on a sphere. A gnomonic projection of this point on a tangent to the sphere at $(\phi_k,\theta_k)$ provides a good projection onto an equirectangular image. Each point on this image will be then a vertex of a graph. Yang et al. \cite{yang2020rotation} and Lv et al. \cite{lv2020salgcn} built regular spherical graphs based on Geodesic Icosaheral Pixelation (GICOPix) and compared an irregular and regular graph to conclude that for a given number of vertices, a regular graph accommodated larger graph isometries. 
\par
Defferrard et al. \cite{defferrard2020deepsphere} utilized a graph representation of the sampled sphere to develop DeepSphere. This model employed graph convolution through a recursive graph Laplacian ($L$). The equivariance condition is $\mathcal{R}(g)Lf = L\mathcal{R}(g)f, \quad \forall g \in \text{SO}(3)$. The score of the faithfulness of the graph representation of the underlying sphere against the normalized equivariance error is $E_L (f,g) = \Bigl (\frac{|| \mathcal{R}(g)Lf - L\mathcal{R}(g)f|| }{||Lf||}\Bigr )^2$. Perraudin et al. \cite{perraudin2019deepsphere} used Hierarchical Equal Area isoLatitude Pixelisation (HEALPix) sampling of the sphere on DeepSphere for cosmological applications. Xu et al. \cite{xu2022hierarchical} used a lifting scheme on hierarchical spherical convolution to learn adaptive spherical wavelets in the model LiftHS-CNN. The lifting enabled a comprehensive hierarchical feature learning via pooling and unpooling. Later, they incorporated local and global attention modules to build global-local attention-based spherical graph convolution (GlasGC) \cite{xu2022rotation}. The computational complexity of a GNN increases with the use of higher-order geometric tensors. However, by embedding the nodes along the edge vectors, Passaro and Zitnick \cite{passaro2023reducing} reduced $\text{SO}(3)$ convolution to equivalent $\text{SO}(2)$ bringing the complexity down from $O(L^6)$ to $O(L^3)$ allowing to scale up geometric models on larger datasets.

\subsubsection{Harmonic Convolutions}

Spherical convolution can be defined as the inner product between a spherical signal $f$ and a rotated filter $\psi$ \cite{cohen2017convolutional, esteves20173d, cohen2018spherical}, both vectors on a sphere $f, \psi: \text{S}^2 \rightarrow \mathbb{R}^K $ resulting in a function on $\text{SO}(3)$
\begin{equation}
    [f \star \psi](R) = \int_{\text{S}^2}\sum_{k=1}^K f_k(x) \psi_k(R^{-1}x) \, dx
\end{equation}
The ensuing layers will employ $\text{SO}(3)$ convolutions $f, \psi: \text{SO}(3) \rightarrow \mathbb{R}^K $, and $R,Q \in \text{SO}(3) $
\begin{equation}
     [f \star \psi](R) = \int_{\text{SO}(3)}\sum_{k=1}^K f_k(Q) \psi_k(R^{-1}Q) \, dQ
\end{equation}
Figure \ref{fig:sphericalConv} illustrates the convolution of a spherical image and a spherical filter, both defined in $\text{S}^2$, yielding an output lifted to $\text{SO}(3)$, followed by convolutions on the group $\text{SO}(3)$. A Generalized Fourier Transform (GFT) \cite{cohen2018spherical} and a related fast algorithm (G-FFT) computed these spherical convolutions. For a manifold $X$ and a Fourier basis $U^l$, the GFT of a function $f : X \rightarrow \mathbb{R}$ is $\hat{f}^l = \int_X f(x) \overline{U^l (x)} \, dx$
\par
This, however, resulted in repeated forward and backward Fourier transforms until Clebsch-Gordan Nets (CGNs) \cite{kondor2018clebsch} generalized this to Fourier Space with a simple implementation. They defined spherical convolution as the inner product of two complex-valued functions, $f(\theta, \phi) = \sum_{l=o}^\infty \sum_{m=-l}^l \hat{f}_m^l Y_l^m (\theta, \phi)$ and $\psi(\theta, \phi) = \sum_{l=o}^\infty \sum_{m=-l}^l \hat{\psi}_m^l Y_l^m (\theta, \phi)$ represented in spherical harmonic expansions indexed by $l=0,1,2,\dots$. and $m \in \{-l,-l+1,\dots,l\}$, where $\hat{f}_m^l Y_l^m= \frac{1}{4 \pi} \int_0^{2\pi}\int_{-\pi}^\pi f(\theta, \phi)Y_l^m (\theta, \phi) \cos \theta \, d\theta \, d\phi$. The spherical cross-correlation for a $R \in \text{SO}(3)$ is then
\begin{equation}
    \left[f \star \psi\right](R)= \frac{1}{4 \pi} \int_0^{2\pi}\int_{-\pi}^\pi f(\theta, \phi) \overline{\left[h_R(\theta, \phi)\right]} \cos \theta \, d\theta \, d\phi
\end{equation}
where $h_R$ is $h$ rotated by $R$, that is $h_R(x) = h(R^{-1}x)$. Then the above integral is simply the outer product
    $\left[\widehat{f \star \psi}\right]_l = \hat{\psi_l} \cdot \hat{f_l}^\dag$,
where $\dag$ is the Hermitian conjugate. Cobb et al. \cite{cobb2020efficient} reduced the greater computational cost of CGN \cite{kondor2018clebsch}, and Spherical-CNN \cite{cohen2018spherical} through constrained generalized convolutions by exploiting the sparsity of CG coefficients. Esteves et al. \cite{esteves2018learning} followed the same approach of Spherical Fourier Transform but with a separation of variables for faster computation using Legendre polynomials $P_m^l$ and a normalization factor $q_m^l$
\begin{equation}
\begin{aligned}
    \hat{f}_m^l Y_l^m & =  \sum_{j=0}^{2b-1} \sum_{k=0}^{2b-1} a_j^{(b)}f(\theta_j, \phi_k) q_m^l P_m^l( \cos \theta_j) e^{-im\phi_k} \\
    & =   q_m^l \sum_{j=0}^{2b-1}  a_j^{(b)} P_m^l( \cos \theta_j)\sum_{k=0}^{2b-1} f(\theta_j, \phi_k) e^{-im\phi_k} 
\end{aligned}
\end{equation}
\par
Liu et al. \cite{liu2018deep} introduced alt-az anisotropic spherical convolution ($\text{a}^3$SConv) with locally-supported filters, offering azimuth rotation equivariance. This convolution is constrained to a set of alt-az rotations, altitude change $\vartheta$, and azimuth change $\varphi$, all evaluated at a point $\hat{\mathbf{u}} \in \text{S}^2$
\begin{equation}
    [f \star \psi](R) = \int_{\text{S}^2}\sum_{k=1}^K f(\hat{\mathbf{u}})(\mathcal{D}_R (\varphi, \vartheta, 0)h)(\hat{\mathbf{u}}) \, ds (\hat{\mathbf{u}})
\end{equation}
where $(\mathcal{D}_R (\varphi, \vartheta, 0)f)(\hat{\mathbf{u}}) = f(R^{-1}\hat{\mathbf{u}})$. An anisotropic spherical CNN using a spin-weighted spherical function (SWSF) represented in spin-weighted spherical harmonics (SWSH) \cite{esteves2020spin} demonstrated exceptional performance with vector field inputs, highlighting its effectiveness as a geometric deep learning architecture. A SWSF $_{s}f: \text{S}^2 \rightarrow \mathbb{C}$ with spin weights $s$, experiences a phase change under $\lambda_\alpha$ rotation by $\alpha$ around polar axis with $\upsilon$ the north pole: $(\lambda_\alpha(_{s}f))(\upsilon) = _{s}f(\upsilon)e^{-is\alpha}$. This phase change restricts SWSHs from transforming among themselves upon rotation. The SWSF is expressed as a linear combination of SWSHs $_{s}Y_m^l$, as shown:
\begin{equation}
   {}_{s}f(\theta, \phi) = \sum_{l \in \mathbb{N}}  \sum_{m=-l}^l {}_{s}Y_m^l(\theta, \phi) {}_{s}\hat{f}_m^l,
\end{equation}
where ${}_{s}\hat{f}_{m}^{l}$ are the expansion coefficients. These coefficients of expansion of an SWSF into the SWSHs transform under group actions, becoming crucial for defining {\itshape equivariant convolutions} between SWSFs as ${}_{s}[ \widehat{f \star \psi} ]_m^l = {}_{s}\hat{f}_m^l {}_{s}\hat{\psi}_m^l$. A $\lambda_g$ rotation for each ${}_{s}f$ by $g\in \text{SO}(3)$, we achieve equivariance: ${}_{s}[\widehat{\lambda_g f \star \psi}]_m^l = \lambda_g[{}_{s}[\widehat{f \star \psi}]_m^l]$.
\par
In addition to the mentioned approaches, recent advancements have led to other independent methods for achieving equivariant spherical representations. For example, Jiang et al. \cite{jiang2019spherical} utilized parameterized Partial Differential Operators (PDOs) to establish spherical convolution on unstructured grids. Additionally, researchers have utilized chart-based rotated parameterized PDOs, as demonstrated in the works by Shen et al. \cite{shen2020pdo, shen2021pdo}, to attain spherical equivariant convolution on a continuous domain. It is sufficient to consider the quotient space, which transforms the principal bundle as seen in the associated bundle to achieve equivariance \cite{cohen2019general}. Building on this insight, Mitchel et al. \cite{mitchel2022mobius} introduced M\"obius equivariant spherical convolution, a spatial aggregation technique. Wavelet-based scattering networks on the sphere utilizing scattering representations \cite{mcewen2021scattering, saydjari2022equivariant} attempt to encode the symmetries in frequency domain rather than physical domain. These methods offer stability to diffeomorphisms across a wide range of frequencies. Bonev et al. \cite{bonev2023spherical} proposed Spherical Fourier Neural Operators (SFNO) that overcame the visual and spectral artefacts caused by Discrete Fourier Transform (DFT). Furthermore, Helwig et al. \cite{helwig2023group} introduced group equivariant Fourier neural operators ($G$-FNO).

\subsection{Scale Symmetries}

Most geometric deep learning models have a multiscale resolution inherent to them that has yet to be fully explained. Scale transformations can appear due to the changing distances between the camera and the object or among the objects. Scale averaging, scale selection, and augmentation are common methods used to address issues related to scale. Exploiting the scale symmetries of the data to achieve a spatial-scale equivariance along with other transformations can enhance the discriminative power of deep learning models.
\par
Deep Scale Spaces \cite{worrall2019deep} can construct scale equivariant convolution using group convolution and the semigroup property: $f(s+t,\cdot) = G(\cdot,s) \ast f(t, \cdot), \forall s,t >0$. The scale-space is Gaussian and constructed from a Gauss-Weierstrass kernel $G(x,t)$. A scale action $S_{A,z}^{\sum_0}$ for dilation $A$, shift $z$, zero-scale covariance matrix $\sum_0$ is defined as $S_{A,z}^{\sum_0}(f)(x) = \left[G_A^{\sum_0} \ast f\right](A^{-1}x+z) $ where $(A,z)$ is the semigroup $S$ element tuples and $G_A^{\sum_0}$ is an anisotropic discrete Gaussian. The semigroup action representation can then be plugged into the group convolution definition for scale-equivariant convolution for discrete scale factors as follows:
\begin{equation}
    \left[f \star \psi\right](A,z) = \sum_{(B,y) \in S}\psi(B,y)S_{B,y}^{\sum_0}(f)(A,y).
\end{equation}
The extensive research and implementation of scale equivariant models drawn on scale transformation action \cite{sosnovik2021disco, zhan2022scale, sangalli2021scale, sangalli2022scale, zhu2022scalingtranslationequivariant} eventually extended the use of the scale-equivariance in 3D data \cite{wimmer2023scale}.
\par
We have already discussed steerability and the advantages of steerable filters in group convolution. Steerable scale equivariant convolutions \cite{sosnovik2019scale, naderi2020scale, sosnovik2021scale, yang2023rotation} avoiding tensor resizing show encouraging research potential in this regard. The scale transformation $f(s^{-1}x)$ can be an upscale $(s>1)$ or a downscale $(s<1)$. The steerable scale equivariant convolution is then:
\begin{equation}
    [f \star \psi](s) = \int_S f(s^\prime) \psi(s^{-1}s^\prime) \, d\mu(s^\prime).
\end{equation}

\subsection{Symmetries of General Manifolds}

Prior discussions on equivariance dealt with global symmetries of the data/feature representations. For manifolds, which generally lack global symmetry, convolutions must address intrinsic space geometry. Convolutions applied to manifolds are ideally characterized by filters with compact support, directionality, and the ability to transfer seamlessly across various manifold structures. In order to achieve these objectives, it becomes imperative to define local gauge-equivariant convolutions.

\subsubsection{Geodesic Convolutions}

{\itshape Geodesic} is a curve or path following the shortest distance between two points on a manifold. {\itshape Parallel transport} translates information along a manifold, preserving the directionality intrinsically. We can assign a frame or an orthonormal basis to a tangent space $T_pM$ at any point $p$ on a manifold $M$. A vector field at $p\in M$ can be evaluated/expressed in terms of this frame, which will be introduced as {\itshape gauge} in the next section. For two points $p, q \in M$, the logarithm of $p$ with respect $q$ gives the "position" of $p$ in $T_qM$. An exponential map $\exp_p: T_pM \rightarrow M$ parameterizes the local neighbourhood $p \in M$ for the parallel transport $\mathcal{P}_{p \leftarrow q}: T_qM \rightarrow T_pM$ along the geodesic from $q$ to $p$ for any vector $v \in T_qM$. These preliminaries would help one build the notion of convolutions on manifolds.
\par
A manifold convolution entails an inner product of two isometric operations. For a point $p \in M$ we identify $T_pM$, expressed by a local frame, whose origin is shifted to $p$. This translation can be a linear isometry $\tau_p: T_pM \rightarrow M$. Now the convolution with a shifted filter $\psi$ can be interpreted as an inner product between this linear isometry and an exponential map $\exp_p:T_pM \rightarrow M$ of a signal $f$ on $M$ as $[f \star \psi](p) = \langle \exp_pf, \tau_p \psi \rangle$. Geodesic Convolutional Neural Networks (GCNN) \cite{masci2015geodesic} generalised the convolutional networks to non-Euclidean manifolds using a patch operator $D(x)$ defined radially. The convolution of a filter $\psi$ applied to this patch on the manifold is mathematically formulated as:
\begin{equation}\label{eqn:geodesic}
    \left[f \star \psi\right](x) = \sum_{\theta,r}\psi(\theta+\Delta\theta,r)\left(D\left(x\right)f\right)(r,\theta)
\end{equation}
\par
Parallel Transport Convolution (PTC) on Riemannian manifolds \cite{schonsheck2018parallel} replaces the Euclidean translation with a vector field transportation mimicking the shift operation by an exponential map. The PTC centered around a point $x_0$  on the manifold $\mathcal{M}$ for a vector field transportation $\mathcal{P}_{x_0}^x$ is mathematically characterized as
\begin{equation}
    \left[f \star \psi\right](x) = \int_\mathcal{M} f(y) \psi \left(x_0,\, \text{exp}_{x_0}\circ\left(\mathcal{P}_{x_0}^x\right)^{-1}\circ \text{exp}_x^{-1}\left(y \right)\right)\, d_\mathcal{M}y
\end{equation}
Directional functions defined on surfaces have demonstrated the ability to allow rotational information to propagate across neural network layers, as shown by Poulenard and Ovsjanikov \cite{poulenard2018multi}. Surface convolution, which commutes with isometry action and parallel transport, termed as field convolutions \cite{mitchel2021field}, lays the foundation for achieving equivariance. Other methods in this regard involve intrinsic kernel parameterizations \cite{boscaini2016learning, monti2017geometric, sun2020zernet, wiersma2020cnns}.

\subsubsection{Gauge Equivariant Group Convolutions}

A gauge is an invertible linear map $w_p: \mathbb{R}^d \rightarrow T_pM$ enabling us to have local frames on $T_pM$. Then a gauge transformation is a change of frame described by the general linear group called {\itshape structure group} $g_p \in \text{GL}(d, \mathbb{R})$ that results in $w_p \mapsto w_pg_p$ and correspondingly $v \mapsto g_p^{-1} v, \, v \in \mathbb{R}^d$. Hence a gauge transformation would result in the corresponding vector invariant in $T_pM$, i.e. $(w_pg_p)(g_p^{-1} v) = w_pv \in T_pM$. The structure group representation describes the group action of a $d$-dimensional geometric quantity as $\rho: G \rightarrow GL(d, \mathbb{R})$. Then a $\rho$-field is a field $f$ that transforms like $f(p) \mapsto \rho(g_p^{-1})f(p)$.
\par
Cohen et al. \cite{cohen2019general} developed a general theory of equivariant convolutions on homogenous spaces, explaining equivariant maps on spheres using fiber bundles and fields. Fiber bundle parameterize isomorphic spaces by another space associated with a group $G$ relative to which geometrical quantities can be expressed numerically. A gauge equivariant convolution defined for processing signals on icosahedron \cite{cohen2019gauge} showcased its representation power on overlapping local charts. This class of architecture could accommodate scalar fields, feature fields and locally flat spaces, thus proving to be a general class of CNN that utilizes intrinsic geometry.
\par
Convolution on a general manifold requires a matrix-valued kernel $\psi$. For $v \in \mathbb{R}^d$, $\psi(v)$ represents a linear map between input and output fibers (feature space) at $p \in M$. If $q_v = \exp_p w_p(v)$, then the kernel should be multiplied by a feature vector at $p$, which is achieved by parallel-transporting $f(q_v)$ to $p$ denoted by $g_{p \leftarrow q_v}f(q_v), \, g_{p \leftarrow q_v} \in G $. A gauge equivariant convolution for a general field will then look like
\begin{equation}\label{gauge_conv}
    [f \star \psi](p) = \int_{\mathbb{R}^d} \psi(v) \rho(g_{p \leftarrow q_v})f(q_v)\, dv
\end{equation}
The gauge equivariance is then, $ [\psi \star f](p) \mapsto \rho(g_p^{-1}) [\psi \star f](p) $. Kicanaoglu et al. \cite{kicanaoglu2019gauge} applied this for gauge-equivariant spherical CNNs in discrete and continuous settings. De Haan et al. \cite{de2020gauge} extended this to Gauge Equivariant Mesh-CNN via equivariant message-passing using parallel transport of features on meshes. The discrete group convolution in this case is: 
\begin{equation}
    [f \star \psi](p) = \sum_{q \in \mathcal{N}_p} \psi_{\mathcal{N}_p} \rho(g_{q \rightarrow p}) f_q
\end{equation}
\par
Weiler et al. \cite{weiler2021coordinate} argued that the chosen structure group imposes the required gauge equivariance through its isometries devised in terms of fiber bundles. He et al. \cite{he2021gauge} incorporated an attention mechanism into gauge equivariance, while Katsman et al. \cite{katsman2021equivariant} expanded gauge equivariance to normalizing flows. Luo et al. further extended this concept to quantum lattice gauge theories in their works \cite{luo2021gauge, luo2022gauge}. Cortes et al. \cite{cortes2023higher} generalized the gauge equivariant convolution in Equation \ref{gauge_conv} from  first-order to higher orders, improving their representational capacity. For a $k$-th order filter, the convolution is
\begin{equation}
    [f \star \psi^{(k)}](p) = \underbrace{\int_{\mathbb{R}^d} \cdots \int_{\mathbb{R}^d}}_{k \rm\ times} \psi^{(k)}(v_1, \dots , v_k) \biggl( \bigotimes_{i=1}^k\rho(g_{p \leftarrow q_{v_i}})f(q_{v_i}) \biggr) \, dv_1 \dots dv_k
\end{equation}
When the interactions within a receptive field are nonlinearly spread in space, higher-order convolutions yield better parameter efficiency. Aronsson \cite{Aronsson2022} devised a more natural setting for G-CNNs that includes gauge equivariant neural networks using homogenous vector bundles $\mathcal{M}=G/H$, where $G$ is an unimodular Lie group and $H$ a compact subgroup. 

\section{Dataset and Applications}\label{data_app}

We will start by listing some datasets commonly used for benchmarking geometric deep-learning problems. Subsequently, we will explore a few applications of equivariant learning.

\subsection{Datasets}
Table \ref{tab:datasets} summarizes various datasets that contain geometric data or are defined in non-Euclidean domain along with the data type.

\begin{table}[ht]
\centering
  \caption{Common Geometric Datasets}
  \label{tab:datasets}
  \begin{tabular}{clc}
    \toprule
    &Dataset&Data type\\
    \midrule
    \multirow{7}{*}{Social Network Datasets} & Facebook Graph Dataset&graph\\
    &Twitter Dataset&graph\\
    &SNAP &graph\\
    & Reddit Dataset &	graph\\
    &	GitHub Social Network Dataset&	graph\\
    &Epinions Social Network Dataset&graph\\
    &YouTube Social Network Dataset&graph\\
    \midrule
    \multirow{6}{*}{Citation Network Dataset} & Cora&graph\\
    &CiteSeer&graph\\
    &WebKB&graph\\
    & PubMed &	graph\\
    &	DBLP Citation Network &	graph\\
    & Citation-network& graph\\
    \midrule
    \multirow{9}{*}{Cheminformatics \& Bioinformatics Datasets}&PubChem&graph\\
    &QM9&graph\\
    &Tox21&graph\\
    &Protein-Protein Interaction Network&graph\\
    &ESOL&point cloud\\
    &MD17&point cloud\\
    &PDBbind&graph\\
    &OC20&point cloud/mesh\\
    &ANI-1&point cloud\\
    \midrule
    \multirow{3}{*}{Traffic Network Dataset}&OpenStreetMap&graph\\
    &PeMS-D7&graph\\
    &NGSIM&graph\\
    \midrule
    \multirow{9}{*}{3D Shape Dataset}&ModelNet&point cloud/mesh\\
    &ShapeNet&mesh\\
    &Stanford 3D Object Dataset&point cloud/mesh\\
    &S3DIS&point cloud\\
    &Dynamic FAUST&mesh\\
    &Pascal3D+&point cloud/mesh\\
    &ScanNet&point cloud/mesh\\
    &KITTI&point cloud\\
    &ScanObjectNN&point cloud\\
    \midrule
    \multirow{4}{*}{Medical Imaging Dataset}&Human Connectome Project&graph\\
    &Visible Human Project&mesh\\
    &BrainWeb&mesh\\
    &LIDC&point cloud\\
    \midrule
    \multirow{2}{*}{Climate Modelling Dataset}&ERA5&voxels/grids\\
    &ClimateNet&voxels\\
    \bottomrule
  \end{tabular}
\end{table}

\subsection{Applications}

Equivariant deep learning models find diverse applications across robotics, computer vision, drug discovery, and material science. This section highlights practical applications wherein geometric data like point clouds and molecular structures play vital roles in classification and regression tasks.

\subsubsection{Computer Vision}

Computer vision's significant success through CNNs is hindered by their inability to process real-world irregular geometric data. Equivariant techniques address this achieving more discriminative power with less data. Graph Neural Networks (GNN) process graph-structured data to model relationships between nodes and edges. GNNs encode local structure based on geometric information and are useful for 3D shape recognition \cite{yu2022rotationally, kumar2022deviant, wu2023transformation, hamdi2021mvtn} and segmentation \cite{fuchs2020se, diao2022superpixel, lei2023efem, deng2021vector}. Point cloud representations as in PointNet \cite{qi2017pointnet} and PointNet++ \cite{qi2017pointnet++} significantly improve voxelization or projection into regular grids by processing directly on point clouds. Equivariance to higher dimensions bolsters scalability \cite{satorras2021n, chen2021equivariant}. Geometric deep learning shows superior performances in registration \cite{wang2022you, ao2021spinnet, bai2021pointdsc, qin2022geometric}, and pose estimation tasks \cite{li2021leveraging, chen2021equivariant, deng2021vector} using point clouds. 
\par
Classification and segmentation of 3D medical images leverage geometric deep learning. Diffusion MRI with diffusion gradients encodes a 3D {\itshape q-space} apart from the 3D image space. Implementation of equivariant methods \cite{muller2021rotation, hussain2023gauge} improve results with fewer samples. Geometric learning models on 3D CT images can significantly improve the classification accuracy of lung nodule diagnosis \cite{dey2018diagnostic}. Geometry-aware Variational Autoencoders \cite{chadebec2022data} bolster geometric learning on medical imaging classifications against translation equivariance.
\par
Spherical cameras popular in virtual reality (VR), augmented reality (AR) systems, robots, drones, and autonomous vehicles demand a different analysis technique, unlike regular 2D image analysis. Rotation equivariant GCNs \cite{yang2020rotation, defferrard2019deepsphere, cohen2018spherical, mcewen2021scattering, esteves2020spin} excel in spherical data analysis. Models tailored to process mesh data and designed to use implicit functions are extensively used for shape completion \cite{chatzipantazis2022se, wiersma2020cnns, basu2022equivariant}, deformation, and reconstruction. These are essential in tasks related to robotic vision. 

\subsubsection{Dynamic Physical Simulations} \label{subsubsec:DPS}

Deep learning can infer the interactions and dynamics of systems in a physical environment over time, considering their rotation, translation and reflection equivariance \cite{kipf2018neural, brandstetter2021geometric, musaelian2023learning}. Equivariant models show superior performance in the natural science tasks of protein binding, molecule design, and chemical property predictions \cite{fuchs2020se, satorras2021n, reiser2022graph, stark2022equibind, liao2022equiformer, gasteiger2021gemnet} than their non-equivariant counterparts. The challenge with multiple interacting objects is that they are geometrically constrained. Huang et al. \cite{huang2022equivariant}, and Magiera et al. \cite{magiera2020constraint} propose constraint-aware models that can encode the constrained dynamics implicitly within the model. Equivariant transformer architecture for molecular dynamics modelling \cite{tholke2022torchmd, tholke2021equivariant} exhibited superior performance and results. The generative model employing normalizing flows developed by Garcia Satorras et al. \cite{garcia2021n} for 3D molecular structures also demonstrated exceptional performance under equivariant constraints.

\section{Challenges and Future Scope}\label{limit}

Geometric learning, employing equivariance as a concrete inductive bias with greater discriminatory prowess than traditional invariant methods, faces limitations and hurdles. We explore these challenges and identify potential future directions to fill these gaps.

\subsection{Limited Data}

While 2D flattened data in the form of images and videos are now ubiquitous, they lack geometric information about objects, rendering them ineffective for geometric learning. Graphs and graph-based learning have recently gained traction. Nevertheless, the availability of 3D data for real-world problems (e.g., medical imaging) and non-Euclidean manifold data (e.g., LiDAR point clouds, omnidirectional spherical images) remains limited and costly. Therefore, effectively implementing stable geometric learning architectures against global and local perturbations continues to be an active field for researchers.

\subsection{Global and Local Descriptor Trade-off}

Balancing global descriptors, which capture higher-level data representations compactly, and local descriptors, which handle intricate patterns and local variations for robustness against slight perturbations, is crucial in equivariant architectures. Achieving this balance remains challenging in tasks like shape completion \cite{wu2022so} and semantic segmentation \cite{sangalli2022scale}. An approach by Simeonov et al. \cite{simeonov2022neural} proposes category-level descriptors for balanced equivariant representation. Recent methods like implicit functions \cite{ye2022gifs, yang2023neural} and occupancy fields \cite{chatzipantazis2022se} address these geometric tasks.

\subsection{Generalization}

The generalization capability of an equivariant model trained on one domain but applied to an unseen one remains a challenge. Models such as gauge equivariant convolutions \cite{cohen2019gauge, basu2022equivariant, cortes2023higher}, trained on one manifold, struggle to generalize to others and higher-order tensors. Many rotation equivariant models compute spherical functions to map meshes or graphs to spheres \cite{maron2017convolutional, liu2021spherical}, leading to flattening. Exploring how natively spherical data could generalize to current models without discretization \cite{cohen2018spherical, coors2018spherenet, bonev2023spherical}, and establishing an invertible mapping from raw data to spherical functions presents intriguing possibilities. Neural Ordinary Differential Equations (Neural ODEs) \cite{gupta2020neural} offer curvature flow-based mapping for enhancing the discriminative power of universal equivariant models through more geometric quantities, equivariance, and non-linearities \cite{brandstetter2021geometric}.

\subsection{Unsupervised Geometric Learning}

Improving the unsupervised learning of symmetries stands as another area for enhancement. Current models rely on inductive biases involving known symmetries and their impact on input and output feature spaces, limiting their ability to capture unknown symmetries. This realm still requires investigation, particularly within the context of equivariant generative models \cite{desai2022symmetry}. Research in this domain could enable the learning of mathematically indescribable symmetries. The theories of soft equivariance \cite{finzi2021residual} and approximate equivariance \cite{wang2022approximately} hold promise for semi-supervised settings in geometric deep learning.

\subsection{Dynamic Modelling}

Expanding from modelling dynamic physical systems on fixed domains covered in section \ref{subsubsec:DPS}, we can extend this to scenarios without fixed domains where signal structures evolve over time. Models capable of tracking time-varying signal structures find applications in anomaly detection for dynamic networks, 3D shapes, and other higher-order geometric properties. Applications span identifying potential peaks and outliers in vast social, traffic, and financial networks, as well as enhancing predictions for 3D motion capture tasks.

\subsection{Hypercomplex and Geometric Algebras}

Researchers have explored alternative convolutions to accommodate geometric transformations beyond group and steerable convolutions. Quaternion convolutions \cite{zhu2018quaternion} leverage quaternion algebra for 3D orientation tasks, capturing intrinsic and extrinsic information. Quaternion algebra presents a promising domain for future geometric representation. Quantum Neural Networks (QNN) \cite{jeswal2019recent} offer avenues for advancing learning-based classification and regression tasks. Machine learning frameworks founded on hypercomplex algebras including quaternions \cite{zhao2020quaternion, comminiello2019quaternion, zhou2023image}, Clifford Algebras \cite{ruhe2023clifford}, Hyperbolic Quaternions, Tessarines, and Klein Four-Group \cite{vieira2022general, takahashi2021comparison}, hold potential for future machine learning models. Additionally, category theory's broader generalization of group theory \cite{de2020natural}, with its potential for replacing equivariance and group representation with naturality and functors, offers exciting prospects for a new era of geometric deep networks and artificial intelligence.

\section{Summary and Conclusion}\label{conclude}

In this review, we comprehensively reviewed the recent deep-learning techniques employing equivariant convolutions, from the background to the methods, in the graph and manifold domains, and covering different symmetry groups. This study broadens the understanding of geometric equivariant representation learning, which outperforms non-equivariant techniques. The main focus is on the approaches employing regular, steerable or PDE-based representations for developing equivariant feature maps generalizing to convolution operations in non-Euclidean domains.
\par
First, we briefly underscore the significance of symmetry groups as a vital inductive bias for deep learning. We focus on equivariance of feature spaces to symmetric transformations. Steerable representations enhance statistical efficiency but suffer high computational costs. This remains a concern for models employing steerable convolutions and other non-trivial representations. Alternative algebras have been proposed to mitigate this problem. The models that utilize equivariant parameterizations differ by domain, symmetry prior, and convolution method. The equivariant convolution approaches outlined in this survey are classified into three groups, each categorized based on their representations concerning the symmetry group transformations they are intended to handle. Their highly diverse mathematical and theoretical underpinnings have been adequately investigated and described with consistent notations. This relationship between the feature maps and the convolution operation in each of these architectures helps us understand geometric encoding in highly irregular data and achieve better learning accuracy than non-equivariant techniques. The review highlights notable examples under various symmetry groups, adaptable to general manifolds via gauge transformations.
\par
The datasets in Table \ref{tab:datasets} aid benchmarking and offer guidance for researchers new to this field. The challenges and limitations of current equivariant methods presented in this review show that there is still room for significant advances leading to the next generation of equivariant deep learning models. Considering potential future directions, we expect that equivariant representation learning will carve out new frontiers in artificial intelligence and scientific computations. We hope this paper will provide a good reference for the machine learning community and researchers in this discipline.

\bibliographystyle{unsrt}  

\end{document}